\documentclass[letterpaper, 10 pt, conference]{ieeeconf}  

\IEEEoverridecommandlockouts                              
                                                          
\usepackage{graphicx}
\graphicspath{ {./images/} }

\usepackage{caption}
\usepackage{subcaption}

\usepackage[compact]{titlesec}
\titlespacing{\section}{0pt}{2ex}{1ex}
\titlespacing{\subsection}{0pt}{1ex}{0ex}
\titlespacing{\subsubsection}{0pt}{0.5ex}{0ex}

\usepackage{xcolor}

\begin{document}
\pagenumbering{gobble}
%
\title{UGV-UAV Object Geolocation in Unstructured Environments}
%
%
%

\author{\parbox{\linewidth}{\centering David Guttendorf, D.W. Wilson Hamilton, Anne Harris Heckman, Herman Herman, Felix Jonathan,
\\Prasanna Kannappan, Nicholas Mireles, Luis Navarro-Serment, Jean Oh, Wei Pu, Rohan Saxena, Jeff Schneider, Matt Schnur, Carter Tiernan, Trenton Tabor}
\thanks{Authors are with National Robotics Engineering Center, the Robotics Institute of Carnegie Mellon University, Pittsburgh PA, listed in alphabetical order. E-mail: wpu@nrec.ri.cmu.edu}
}

\maketitle

\begin{abstract}
A robotic system of multiple unmanned ground vehicles (UGVs) and unmanned aerial vehicles (UAVs) has the potential for advancing autonomous object geolocation performance. Much research has focused on algorithmic improvements on individual components, such as navigation, motion planning, and perception. In this paper, we present a UGV-UAV object detection and geolocation system, which performs perception, navigation, and planning autonomously in real scale in unstructured environment. We designed novel sensor pods equipped with multispectral (visible, near-infrared, thermal), high resolution (181.6 Mega Pixels), stereo (near-infrared pair), wide field of view (192 degree HFOV) array. We developed a novel on-board software-hardware architecture to process the high volume sensor data in real-time, and we built a custom AI subsystem composed of detection, tracking, navigation, and planning for autonomous objects geolocation in real-time. 

This research is the first real scale demonstration of such high speed data processing capability. Our novel modular sensor pod can boost relevant computer vision and machine learning research. Our novel hardware-software architecture is a solid foundation for system-level and component-level research. Our system is validated through data-driven offline tests as well as a series of field tests in unstructured environments. We present quantitative results as well as discussions on key robotic system level challenges which manifest when we build and test the system. This system is the first step toward a UGV-UAV cooperative reconnaissance system in the future.  
\end{abstract}


%
\IEEEpeerreviewmaketitle

\section{Introduction}
\begin{figure}[t]
\centering \includegraphics[width=0.4\textwidth]{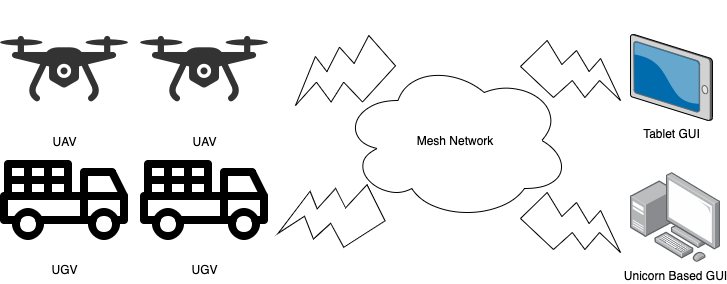}
\setlength{\belowcaptionskip}{-14pt}     
\caption{Illustration of the system. UAVs and UGVs detect objects of relevance and send their locations to human operators through wireless mesh networks.}
\label{fig:sys_illustration}
\end{figure}

Recent advances in computer vision, machine learning, autonomous planning and navigation, and human-machine interfaces have made highly functional robotic applications possible.  In spite of the progress, designing and implementing such a full system to achieve end-to-end real world objectives remains a daunting research challenge.
In this work, we present our design, implementation, and demonstration of one such such system: a multiple UGV-UAV object geolocation system. The robots used in the system are equipped with our custom, long range, high-resolution, multispectral camera arrays. With these, the robots are capable of performing perception, navigation, and motion planning autonomously. As illustrated in Fig.\ref{fig:sys_illustration}, the geolocations of objects detected by individual robots are transmitted to GUIs to form a Common Operating Picture (COP) for human operators via a wireless mesh network. The system is designed for off-road unstructured environments although the design could be adapted for structured environments as well. We will describe the on-board hardware-software architecture and the new sensor pod. We customized state-of-the-art AI functionalities including perception, tracking, navigation, and planning to demonstrate the capability of the new system. Although the system now provides a great testbed for it, algorithmic research on individual AI components is beyond the scope of this paper.  

We designed the system to meet the following high level goals and constraints: 1) it must reliably detect and localize far away (hundreds of meters) objects under a variety of lighting conditions. Therefore, high resolution cameras working at different spectra with wide field of view (FOV) are necessary; 2) some locations in the working area are not traversable, even for all-terrain vehicles, and some of the objects are occluded from a bird's eye view (e.g. under tall trees with dense foliage); 3) the robots in the system should work fully autonomously and send the locations of relevant objects in real-time through a wireless mesh network to an interface where human operators can see them on a map; 4) passive sensing for object detection is preferred over active sensing; 5) limits on communication bandwidth and reliability preclude off-board sensor processing and control as well as centralized planning and coordination.

To the best of our knowledge, this is the first prototype reported in the literature which satisfies all of the above requirements. The contributions of this paper are summarized as follows:
\begin{enumerate}
    \item We designed and implemented a robotic system to meet the goals listed above.  
    \item We designed and fabricated a novel semi-modular sensor pod to achieve a  state-of-the-art combination of sensing capabilities: a) Total resolution: By combining 5 sensor modules into one unified sensor pod, it is able to capture 60 Mega Pixels in the visible light (RGB) spectrum, 120 Mega Pixels in near infra red (NIR) spectrum, and 1.6 Mega Pixel in long-wave infrared a.k.a. thermal (LWIR) spectrum in single snapshot. b) Field of view: The sensor pod has a  $36^{\circ}$ vertical FOV and $192^{\circ}$ horizontal FOV. c) Multi-spectrum: Each sensor module can perceive the environment through RGB, NIR, LWIR.
    \item We demonstrated the resulting system in multiple field trials yielding both a qualitative evaluation and a quantitative analysis of its system level performance for the stated goals.
\end{enumerate}

Self-driving cars are another highly functional robotic system with difficult end-to-end performance metrics and they share some of the same components as ours. However, unlike most self-driving cars our system is a multi-robot system, operates in off-road environments, and uses only passive sensing.  This results in the unique set of challenges, solutions, and opportunities for future systems level research that we present here.


\section{Related Work} \label{sec:rw}
\subsection{Multispectral Sensing}
An RGB camera is sensitive to lighting conditions. 
In this work, we build a unified sensor pod packing RGB, LWIR, and NIR sensors together. A related sensor pod is presented in \cite{thermal_sensor}, albeit with a very different design focus. In \cite{thermal_sensor}, the authors design a multispectral cost effective sensor pod mainly for indoor applications; while we design and build a rugged sensor pod for outdoor mobile robotic applications. For example, \cite{thermal_sensor}'s working depth range is 0.35 - 1.4m, while our system's working range needs to exceed 150 meters. Similarly, \cite{tokyo_multispectral} employs an array of RGB and IR cameras to collect a multispectral dataset at 1 Hz. While the authors employ only front-facing cameras at a limited resolution of maximum 640x480, our novel sensor pod has a FOV of $192^{\circ}$ x $36^{\circ}$ while processing high-resolution (a maximum of 4096$\times$3000) multispectral images at 4 Hz.

\subsection{UGV and UAV Reconnaissance}
There is a history of using multiple UGVs and UAVs together to accomplish certain reconnaissance missions. In \cite{Uav19}, a UAV attaches a tether atop a hill and a UGV climbs by winding it, thus enhancing the traversability of the UGV. They demonstrate their system using a miniature UAV and UGV in an indoor lab environment. 
\cite{petrovic2015can} and \cite{tanner2007switched} also describe heterogeneous UAV-UGV systems which are tested solely in simulation.  
On the other hand, in \cite{Balta2020}, the authors propose a robotic system for registering and segmenting 3D point clouds of large-scale natural terrain using geometry based methods. Our work is different from these systems in that we focus on full-sized vehicle platforms in real off-road environments with complete AI capability, from object geolocation to autonomous navigation and motion planning.

In \cite{surmann20173d}, \cite{kim2019uav}, the authors design UAV-UGV systems to map 3D environments. While their systems (especially UGVs) rely heavily on LIDAR to build these maps, we employ primarily passive and multispectral sensors to localize relevant objects on a map.

\section{System Architecture} \label{sec:sa}
In this section, we discuss sensor pod design, the UGV and UAV platform, as well as the associated on-board system software stack. A discussion of the various AI subsystems can be found in the next section.

\subsection{Sensor Pod Design}
\begin{figure}[t]
    \centering
     \begin{subfigure}{0.10\textwidth}
         \centering\includegraphics[width=\textwidth]{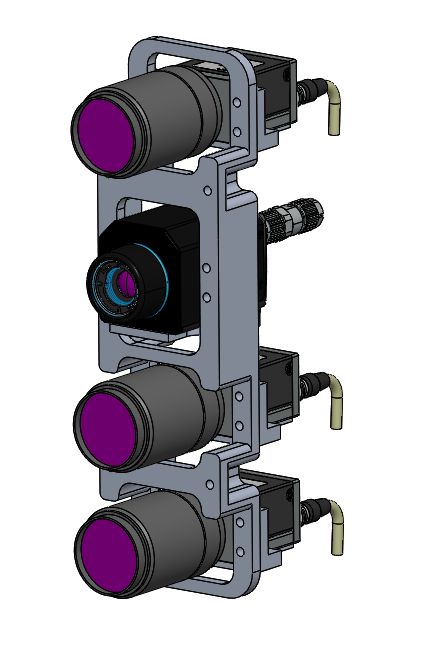}
     \end{subfigure}
     \begin{subfigure}{0.10\textwidth}
         \centering\includegraphics[width=\textwidth]{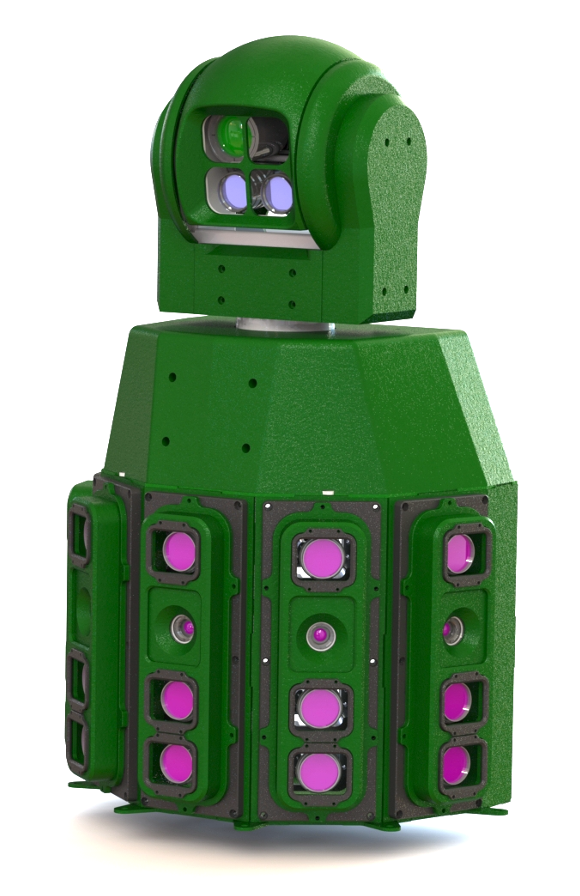}
     \end{subfigure}
     \begin{subfigure}{0.16\textwidth}
         \centering\includegraphics[width=\textwidth]{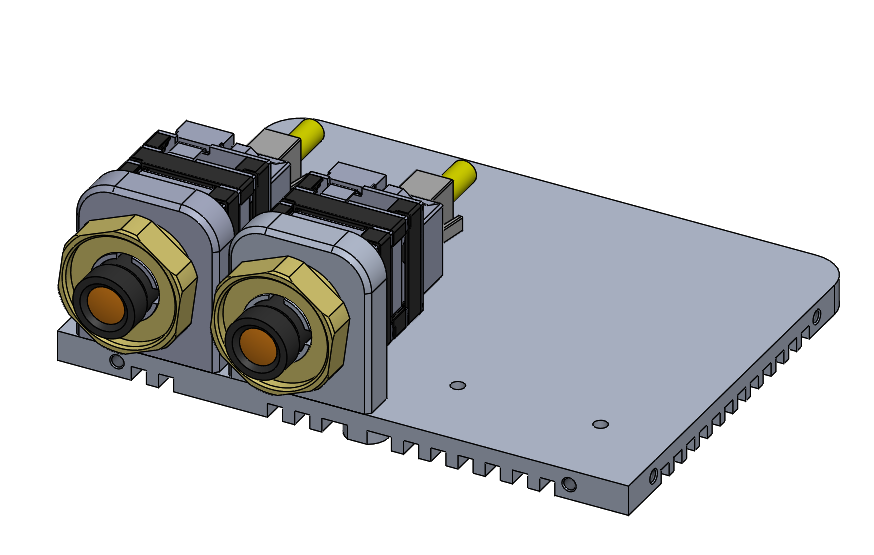}
     \end{subfigure}
\setlength{\belowcaptionskip}{-14pt} 
\caption{Left: UGV sensor module. Top down: NIR 1, thermal, RGB, NIR 2.
Middle: UGV sensor pod, composed of 5 sensor modules and 1 pan-tilt camera module. Right: UAV sensor pod. One RGB, one NIR, and one slot for future expansion.}
\label{fig:sensor_module}
\end{figure}

\begin{figure}[t]
    \centering
     \begin{subfigure}{0.20\textwidth}
         \centering\includegraphics[width=\textwidth]{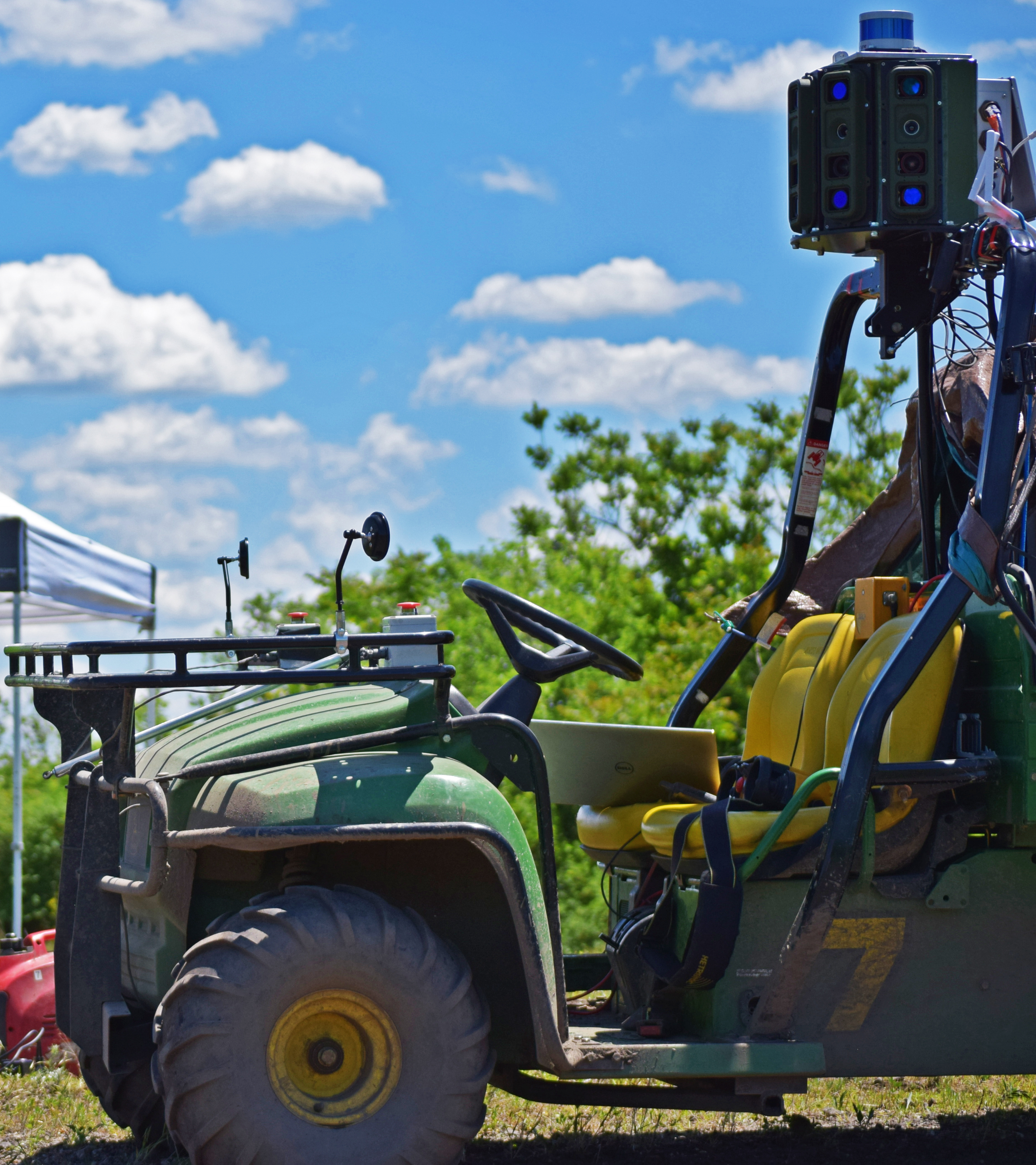}
     \end{subfigure}
     \hfil
     \begin{subfigure}{0.215\textwidth}
         \centering\includegraphics[width=\textwidth]{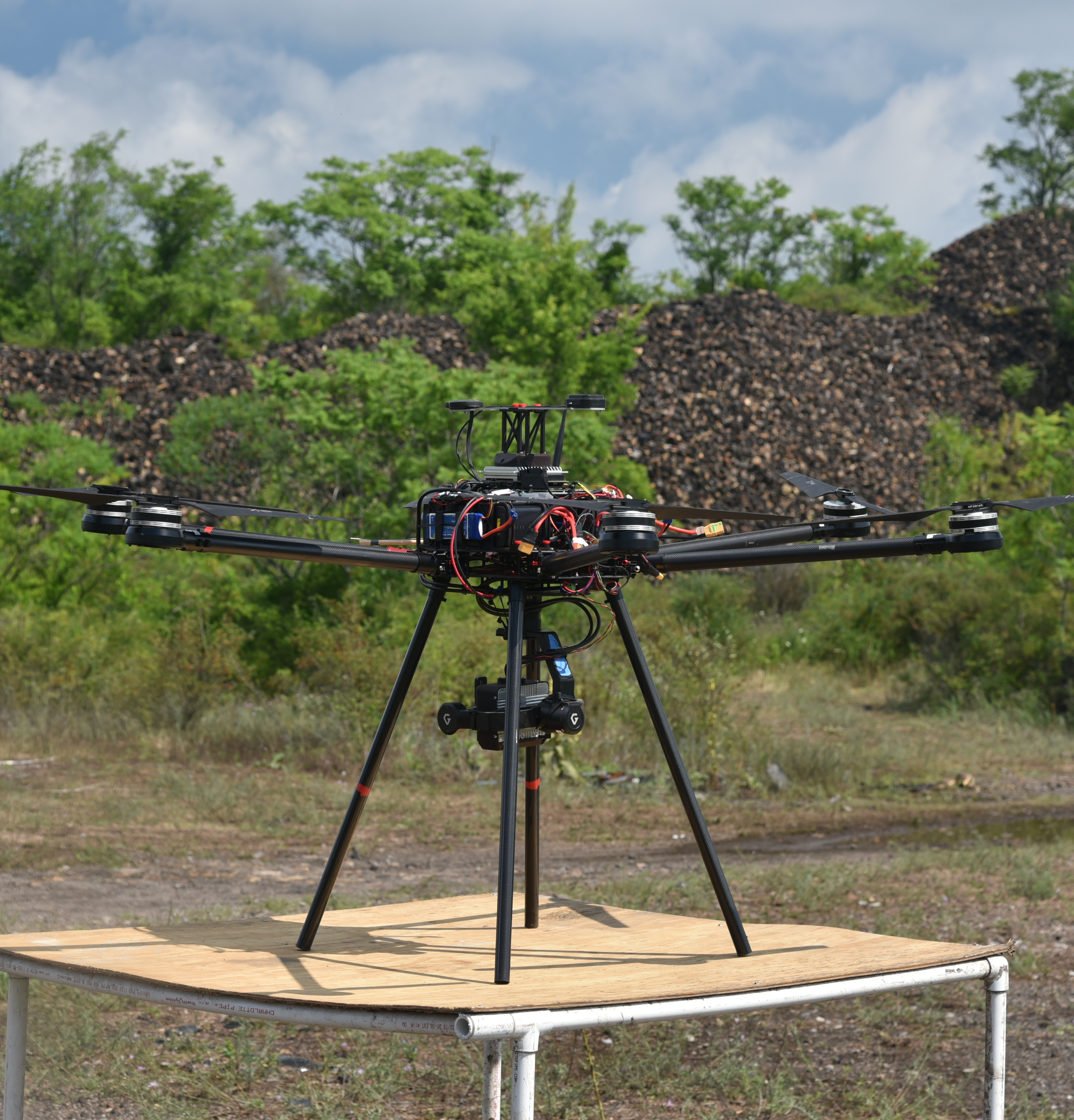}
     \end{subfigure}
\setlength{\belowcaptionskip}{-14pt}
\caption{Left: RecBot UGV. Right: Lil Hexy UAV.}
\label{fig:gator_pod}
\end{figure}

We designed a semi-modular sensor pod to allow different configurations of sensing. An individual module is illustrated in Fig. \ref{fig:sensor_module} left. Multiple modules can be installed side-by-side to expand the field of view (FOV). For example, in our UGV platform (Fig. \ref{fig:gator_pod}), we combined 5 modules to form a $4\times5$ camera array in order to cover a horizontal FOV of $192^{\circ}$ and a vertical FOV of $36^{\circ}$.  In this configuration, adjacent modules have an overlap of $12^{\circ}$ in the RGB and NIR spectrum and $9^{\circ}$ in thermal (LWIR).

The specification of the sensor pod is as follow:

\begin{itemize}
\item FLIR A65 Thermal camera: a 640x480 pixel commercial off-the-shelf (COTS) thermal camera to cover a horizontal FOV of $45^{\circ}$ and a vertical FOV of $36^{\circ}$.

\item RGB Camera: a 4096$\times$3000 pixel (12 MP) RGB camera to cover a horizontal FOV of $48^{\circ}$ and a vertical FOV of $36^{\circ}$.

\item NIR Stereo Pair: two 4096$\times$3000 pixel monochrome cameras with a NIR longpass filter to cover a horizontal FOV of $48^{\circ}$ and a vertical FOV of $36^{\circ}$ with a baseline length of 0.2 meters.

\end{itemize}

In addition to the semi-modular sensor pod, the system also incorporates a Pan-Tilt module.  The Pan-Tilt system uses a similar camera payload as the modules in the semi-modular omitting one of the NIR cameras and adding a laser range finder.  The lenses selected for the Pan-Tilt module provide a horizontal FOV of $16^{\circ}$ and a vertical FOV of $12^{\circ}$.  The purpose of the pan-tilt module is to provide a closer look at any potential detection.

A sensor pod with one RGB camera and one NIR camera is deployed on the UAV. They are COTS cameras with 2048$\times$1536 pixel (3.2 MP) to cover a horizontal FOV of $60^{\circ}$ and a vertical FOV of $45^{\circ}$.

\vspace{-4pt}
\subsection{Vehicle Platforms}
For the UGV we use the RecBot, a John Deere E-Gator utility vehicle that has been retro-fitted with a drive-by-wire autonomy kit. The RecBot supports receiving velocity and curvature commands as well as providing actual velocity and curvature as feedback. The RecBot also supports e-stops, both physical and software. The RecBot allows a safety driver to be on the UGV to take over in an event of an emergency. A gasoline powered generator was added to the E-Gator to power the sensors and the computing platform. The output of the generator was fed to a battery backup before powering the compute. This battery backup, in addition to providing a few minutes of battery power to the compute, conditioned the power to the compute. Circuitry was designed to allow cycling of power to each of the sensors individually as past projects have shown this is an invaluable way to recover non-responsive sensors and make the system more robust. The intended operation time of a UGV is up-to 4 hours.

For the UAV we used the Lil Hexy \cite{hexy} air-frame paired with a PixHawk flight controller \cite{pixhawk} as our base platform.  To this we added a custom camera payload and additional on board compute in the form of the Nvidia Jetson AGX Xavier. The intended operation time of a UAV is up-to 45 minutes.

\begin{figure}[b]
\centering\includegraphics[width=0.47\textwidth]{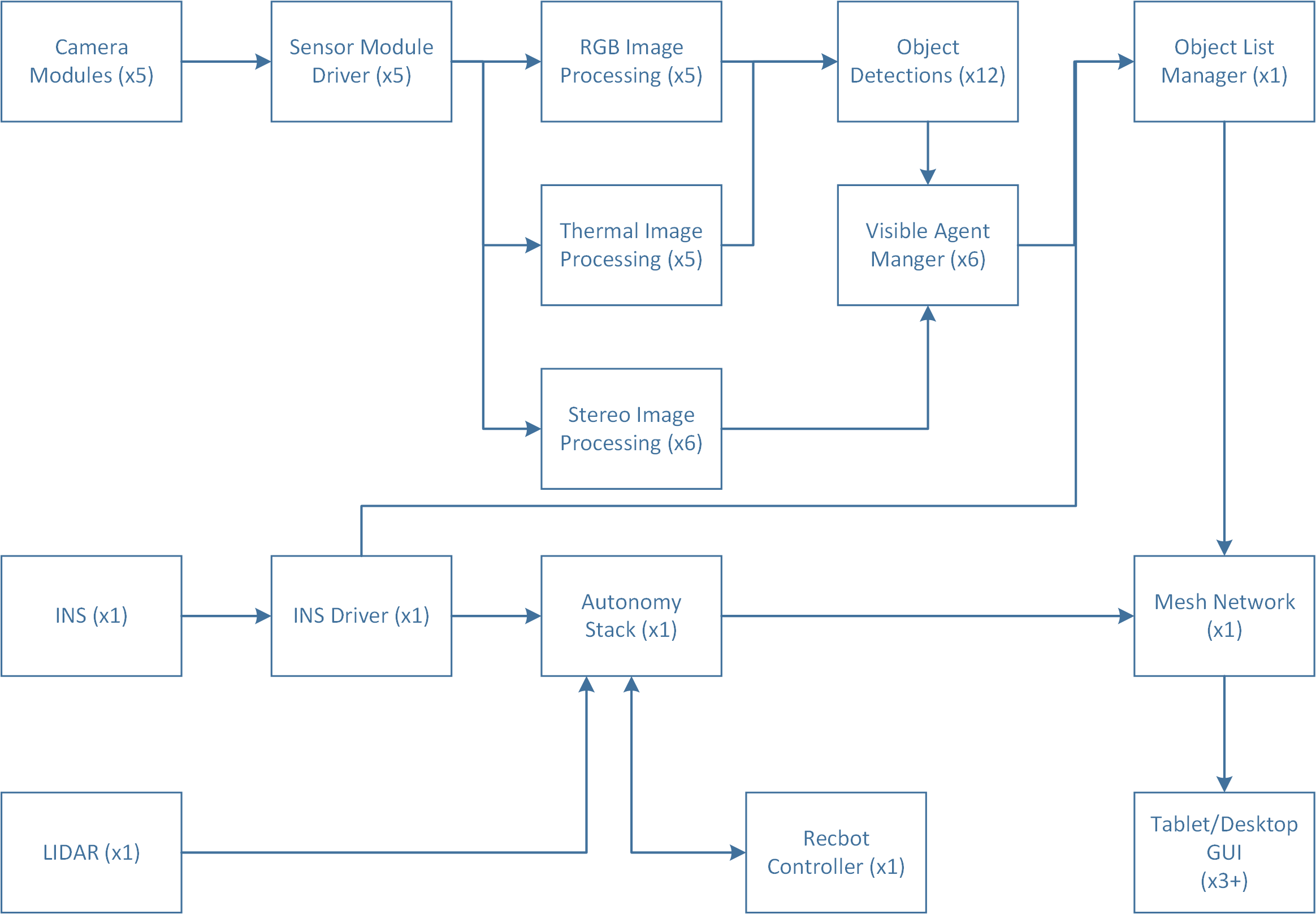}
\caption{On-board software stack}
\end{figure}

\subsection{On-board Hardware/Software Codesign}
The UAV and UGV share the same hardware-software design philosophy. Due to page limit, we focus on the UGV and only briefly discuss the UAV components which are substantially different from their UGV counterpart.

\begin{figure}[t]
\centering\includegraphics[width=0.47\textwidth]{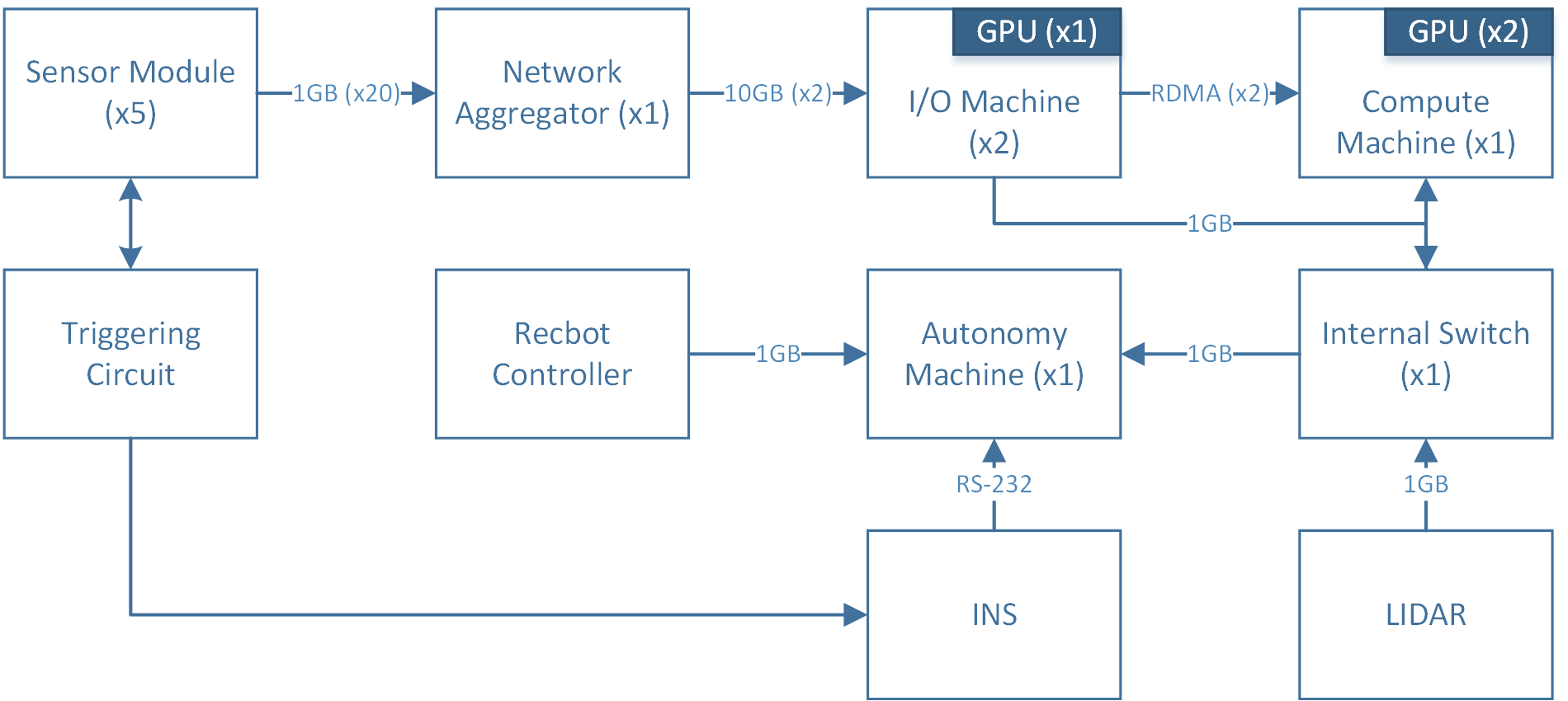}
\setlength{\belowcaptionskip}{-14pt}
\caption{On-board computer configuration}
\label{fig:onbord_compute}
\end{figure}

Due to the large amount of data coming from the sensor pod (approx. 728 MPixels per second), the compute for the UGV was divided into 4 pieces, two I/O machines, one compute machine, and one autonomy computer. The two I/O machines handle up to 3 sensor modules each. They pre-process the images streaming from the cameras and use the GPU to calculate a point cloud for each module using the two monochrome cameras in each sensor pod. The pre-processed images and point clouds are transferred to the compute machine for the perception module to further process as described below. The transfer between the I/O machines and compute machine is utilizing Remote Direct Memory Access (RDMA) which greatly reduces the latency and CPU overhead on the machines for handling the network transfers. The autonomy computer interfaces with the LIDAR, Initial Navigation System (INS), and RecBot controller and currently runs the entire autonomy stack. INS data is forwarded to the compute machine from the autonomy machine over a TCP link. 

A triggering circuit was designed to ensure all the cameras are capturing images at the same time, in addition this triggering circuit is an input to the INS. This allows us to synchronize the clocks between the cameras and the INS so the pose of the robot could be associated with each image accurately. In addition, all machines and non-thermal cameras are utilizing the Precision-Time-Protocol (PTP) with one of the I/O machines as the master to keep their clocks in sync. 

The machines are utilizing a in-house designed and implemented robotics infrastructure, called Unicorn, for health and status monitoring, message passing, serialization, logging, configuration, and various other utilities commonly used on robotics programs. This Unicorn infrastructure is designed with throughput in mind and utilizes zero-copy message passing and de-serialization where possible. Unicorn cooperates with Robotic Operating System (ROS) through an optimized Unicorn to ROS message translator.

The Autonomy Stack currently utilizes the ROS to facilitate re-using components from previous research projects. 

The hardware and software on the UAV follow the same structure as that on the UGV, but the specifics have been significantly modified in order to respect the weight limit of the UAV platform. The Nvidia Jetson AGX Xavier onboard the UAV handles the responsibilities of the I/O machines and compute machine used on the UGV, while the Pixhawk flight controller handles the responsibilities of the autonomy machine. Similar to the UGV, the Nvidia Jetson AGX Xavier runs software that utilizes both the Unicorn infrastructure and ROS, controls a triggering circuit to synchronize the capture of data from its cameras and INS, and handles preprocessing images and performing object detection on them.

\section{AI Subsystem} \label{sec:ai}

\subsection{Perception}
The perception module consumes the raw sensor data and produces object tracks in global coordinate system. It contains three main functions: 2D object detector, depth estimator, and tracker. The detector locates objects' bounding boxes on input images while in parallel, the depth estimator generates point cloud with respect to camera coordinate so that 2D bounding boxes can be mapped into 3D space. Then localization information from navigation module is used to transform them to global coordinate. The tracker filters objects' locations in global coordinate, rejecting false positive detections and making predictions when there are false negative detections. Both detection thumbnails and track positions are streamed to remote control center through mobile ad-hoc networks for use by human operators.

\subsubsection{Object Detection}
The detector detects and classifies objects of relevance to the task. For UGV detector, the computing power is abundant. We customize the widely used YOLOv3 neural network architecture \cite{yolov3} to support fast inference on Nvidia RTX 6000 GPU given very high resolution input images. Specifically, the \textit{darknet} trunk is trimmed for speedup, the number of object classes are replaced to those relevant to our task, and we optimize the deep learning model's IO performance. We only use RGB images as input to the detector for now. We are working on building a multispectral version to fully utilize our sensor pod's capability. For UAV detector, the computing power is limited, we use a less computation intensive model \cite{multianchor} and run it on Nvidia Jetson AGX Xavier.

Due to the significant domain shift between our off-road environment and popular 2D object detection public datasets, we collected our own dataset and developed the associated data management tool infrastructure (Sec. \ref{sec:data}).

\subsubsection{Stereo Estimation}
To support the high resolution input images, we adopt the fast stereo depth estimator proposed in \cite{yang2019hsm}. To train the stereo model, we first generate data from CARLAR simulator \cite{Dosovitskiy17} using scenes from 5 towns. In the simulator, we create a sensor pod with the same lens layout as our real sensor pod. We pretrain our model with this generated dataset whose disparity ground truth is available.

We then fine tune the model to fill the domain gap between simulated data and real data. Labeling disparity ground truth is hard. 
We tackle this issue from two approaches. First, we create pseudo labels for our own dataset by applying piecewise rigid scene model (PRSM) \cite{prsm}, a state-of-the-art classical computer vision method. The model is fine tuned using supervision from these pseudo ground truth. Second, we apply self supervised learning using multi-scale photo-metric loss \cite{unsupervise}. This loss encourages local pixel level consistency between the stereo pairs after warping with respect to the predicted disparity. 

\subsubsection{Global Tracking}
In order to map objects in space, we combine detections in a modified Kalman filter tracker, similar to \cite{chiu2020probabilistic}. In our application, we assume that the objects are static. We make two adaptions. Our first change is in how we parameterize the Kalman filter. Because we are trying to build object maps and currently assuming that objects are static, we use a constant position model and we set the process noise to 0. The second change is in the birth/death of tracks. Since our robot is operating in the same area for an extended period of time, we would like to match detections across a longer time horizon. We therefore only let a track die if it does not receive any further update for a relatively long period of time. As detections can be intermittent due to false negatives, we generalize the birth rule from \cite{chiu2020probabilistic} and require that a track must have at least $N$ (default 3) matches where any period of no detection in between does not span longer than $t$ (default 30) seconds.

In \cite{chiu2020probabilistic}, 3D detections were provided as the inputs to the tracker. While this is easy in a LIDAR based detector, it was necessary for our system to merge object detection and stereo results.  We first take all of the points from stereo that fall within the detected bounding box after 3D-2D projection. 
If there are enough stereo points in the bounding box, we take the median depth. If there are not enough points or the median depth is too far for stereo to be accurate, we use a default depth value. 

This prior work \cite{chiu2020probabilistic} used a constant, diagonal covariance matrix for observations. 
In our system, we know that we have more uncertainty in the \textit{range} to an object than in the bearing to it. 
We model this with a diagonal covariance aligned with the bearing to the object, where each of the diagonal elements is proportional to the square of range. The coefficient on the two bearing related directions are identical and the coefficient on the range direction is much larger. We then rotate this covariance into the global frame (UTM) for matching to existing tracks. This rotation results in an observation covariance that is not diagonal, better modeling the uncertainty of our sensors. Currently we only run tracking in UGVs. 

\subsection{Navigation}
The navigation system is responsible for controlling the robotic platforms to achieve a sequence of input way-points while localizing itself and reacting to different features in the vehicle's environments. Multiple UAVs and UGVs platforms are tasked here with different mission objectives in form of way-point sequences. The state of the environment as sensed by each of these platforms is combined to generate the COP. A finer understanding of the navigational capability of a platform can be realized by looking at the planning algorithms, localization sub-system and perception inputs utilized by the navigation system. 

Most of the details in the following sections are geared towards the UGV platforms. The UAV platforms we use are holonomic, and we assume that there are no obstacles at the altitude at which our UAV platforms operate. These considerations greatly simplify the planning problem problem for UAV platforms. The planning and control is handled using the on-board ArduCopter \cite{arducopter} multicopter UAV controller software, which runs on the Pixhawk flight controller hardware.

\subsubsection{Localization}

Localization is performed at two levels, namely, low-frequency discrete global pose measurements and high-frequency continuous local pose measurements. GPS measurements with RTK augmentation provided 6-DOF global pose at centimeter level accuracy. To generate local pose estimates, wheel odometry and inertial measurements from the on-board INS unit are fused to generate measurements at 50 Hz.

\subsubsection{Navigation Cost-maps} \label{costmap}

For navigation, the environment is represented using a global cost-map and a local cost-map. The former is initialized from prior information registered in a global coordinate frame, and is used by the global planner. The latter is generated from sensor data and registered in a local coordinate frame as the robot traverses the environment, and is used by the local planner. The global cost-map used in our experiments covers the total extent of the test area, and it was obtained from a point cloud model acquired using airborne 3D LIDAR scanners. From this model, navigable paths were identified using a geometric analysis over regions of points. The terrain roughness was measured by first fitting a plane to all the points within a radius of 1.5m, and then computing the average distance of each point inside the region to the plane. Finally, the navigable areas are demarcated by applying a threshold to the terrain roughness for each cell in the map, and projected onto a 2D cost-map, with a cell size of 0.5m. The threshold value was determined empirically by verifying that the resulting cost-map accurately represents the network of unpaved roads present in the test area.

The local cost-map is generated at run time: first, LIDAR measurements from a single scan cycle are filtered based on their distance to the sensor, and then voxelized to further reduce the number of points to process. To allow the system to operate in the irregular surfaces commonly found in outdoor environments, the space is partitioned as a grid, for which the ground elevation of each cell is estimated. The mean and standard deviation of the heights of all points inside each cell are computed recursively, and the cell elevation is calculated by subtracting one standard deviation from the mean. This results in elevation that is never below the lowest point while still having about 80\% of points above ground. The ground elevation is stored in a scrolling grid that is centered 15m ahead of the robot’s current position. The ratio of the number of points located above ground and the total number of points inside the cell is calculated and used to estimate the probability of a cell being occupied by an obstacle. 

\subsection{Planning}
\subsubsection{Global Planning} \label{global_planning}

In order to traverse a sequence of way-points, a global planner generates a trajectory that connects the sequence of input waypoints while accounting for the constraints imposed by the global map prior and dynamics of the vehicle. The ARA* planner \cite{likhachev2004ara}, chosen here, allows to accommodate the Ackermann steering constraints of UGV and provides for a fast, albeit sub-optimal, trajectory search across the global cost-map. The Ackermann steering constraints are incorporated into the planning problem by choosing motion primitives that satisfy the UGV's dynamics. The solution to this planning problem is a dense sequence of points representing a trajectory that passes through all the input way-points. This output trajectory is fed to the local planner for actuating the UGV and ensuring that this trajectory is followed.

\subsubsection{Local Planning}
The act of following a planned trajectory in the presence of noisy localization inputs and detected environmental features, possibly obstacles, is the responsibility of the local planner. The local planner uses the traversibility information from the local cost-map to generate plans that will help the robot stay close to the trajectory generated by the global planner while simultaneously avoiding obstacles in the robot's environment. A implementation of pure-pursuit local planner \cite{coulter1992implementation, snider2009automatic} was used to track the trajectory generated by the global planner.

\section{Data Pipeline And Data Management} \label{sec:data}

\subsection{Data Annotation}
Raw data logs collected in the field are first manually segmented to remove any segments of the data that was accidentally recorded, redundant, or otherwise unusable. At this stage, the segments are also annotated with the type of labeling to be done. After the data is segmented, an output \emph{JSON} file is generated containing the timestamps of the newly created segments, respective to the original log, and all their associated metadata. This file is then parsed by a recurring job and the information placed into the database. These segments form the basis for the subsequent workflow.

Unlabeled data segments in the database are uploaded to a cloud-based annotation tool and a corresponding task is created within the \emph{Atlassian Jira} ticketing system. Annotators claim and complete available tasks. Completed tasks are verified to meet quality standards by a different individual before the annotations are downloaded and registered to the database. Annotator performance and overall progress is tracked automatically via relations within database linking annotators and completion dates to annotations. These metrics are visualized using custom \emph{Grafana} dashboards.

\subsection{Dataset Generation}
In support of the aforementioned models, unbiased, evenly-distributed data is required. To this end, we have developed a small Python utility to generate collections of annotated data segments. These collections consist of the traditional training, validation, and test sets. In this interactive utility, users define the percentage of data segments allocated to each of these splits. 
After these splits are defined, data segments are allocated such that the various categories of metadata are evenly distributed across all three splits as evenly as possible.

\section{Human-Robot Teaming} \label{sec:interface}

It is a challenging task for a human operator to command a fleet of robotic vehicles spread over a large area to perform a coordinated task. The objective of designing an interface supporting such human-robot communication and collaboration is to reduce cognitive complexity on the human operator’s end while retaining critical information in the communication channel. 

For the purpose of this work, a tablet-based interface known as Android Team Awareness Kit (ATAK \cite{atak}) has been used. ATAK is a freeware Geographic Information System (GIS) that provides support for geospatial reasoning and situation awareness in various problem domains including military applications. In our experiments, we demonstrated that a human operator assigns a high-level mission to a team of 4 robots, two UGVs and two UAVs. The robots operate in a fully autonomous mode to perform the high-level vigilant navigation tasks that have been assigned by the human operator where the robots are expected to recognize objects of interest along their routes. The pieces of information each robotic vehicle acquired, e.g., objects of interest detected with high confidence, are passed to the interface where the information can be presented to the operator in a hierarchically organized fashion.

\section{Experiments} \label{sec:exp}
We present offline evaluate results for the AI system and field test results for the whole system. 

\begin{figure}[h]
    \centering
    \begin{subfigure}{0.15\textwidth}
        \centering\includegraphics[width=\textwidth]{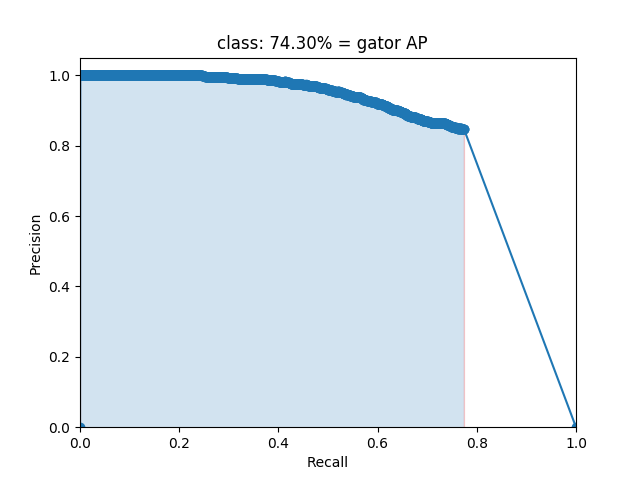}
    \end{subfigure}
    \begin{subfigure}{0.15\textwidth}
        \centering\includegraphics[width=\textwidth]{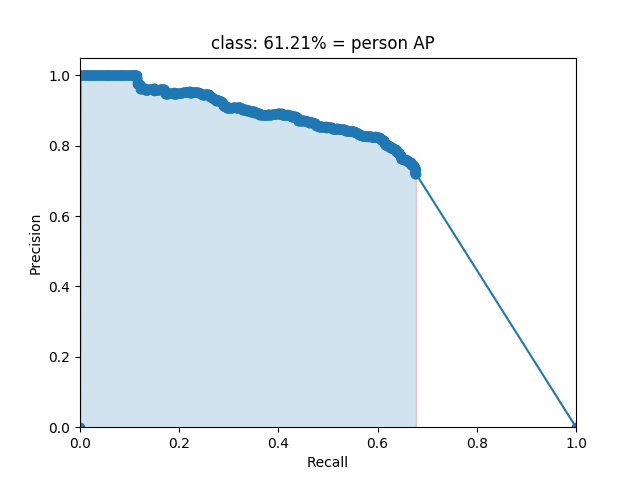}
    \end{subfigure}
    \begin{subfigure}{0.15\textwidth}
        \centering\includegraphics[width=\textwidth]{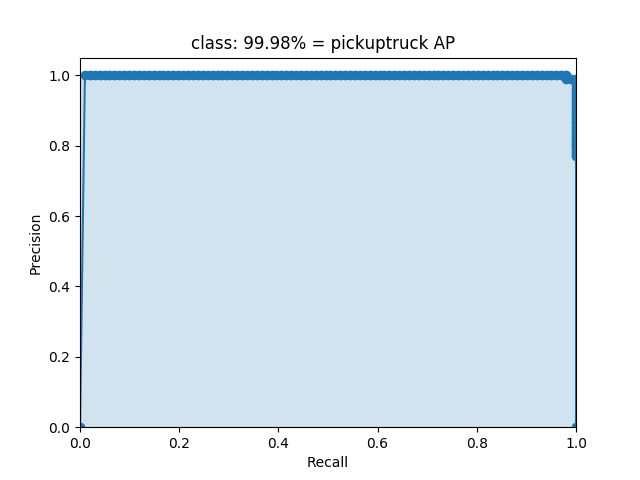}
    \end{subfigure}
    \setlength{\belowcaptionskip}{-14pt}
    \caption{Precision-Recall curves of UGV detector.}
    \label{fig:gatormap}
\end{figure}

\begin{figure}[h]
\centering\includegraphics[width=0.4\textwidth]{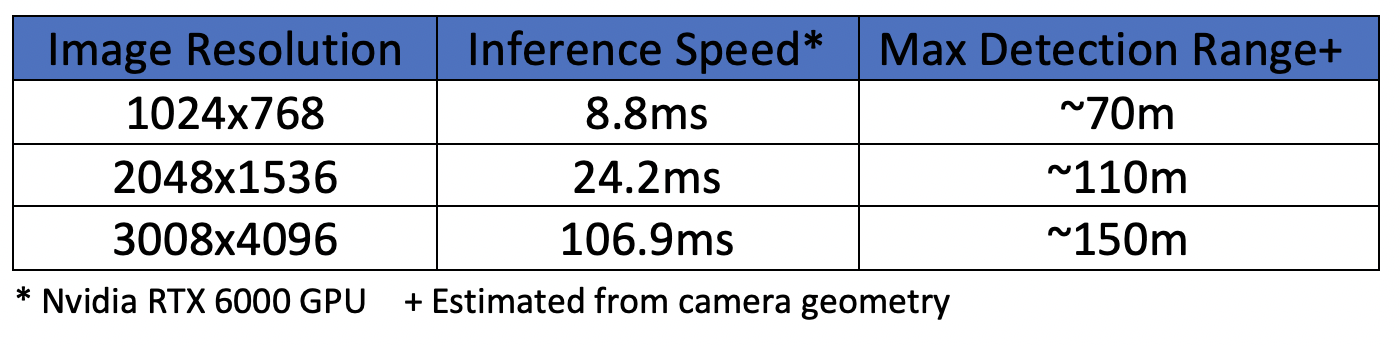}
\setlength{\belowcaptionskip}{-14pt}
\caption{UGV 2D detector benchmark.}
\label{fig:detbenchmark}
\end{figure}


\begin{figure}[b]
     \centering
     \begin{subfigure}{0.15\textwidth}
         \centering
         \includegraphics[width=\textwidth]{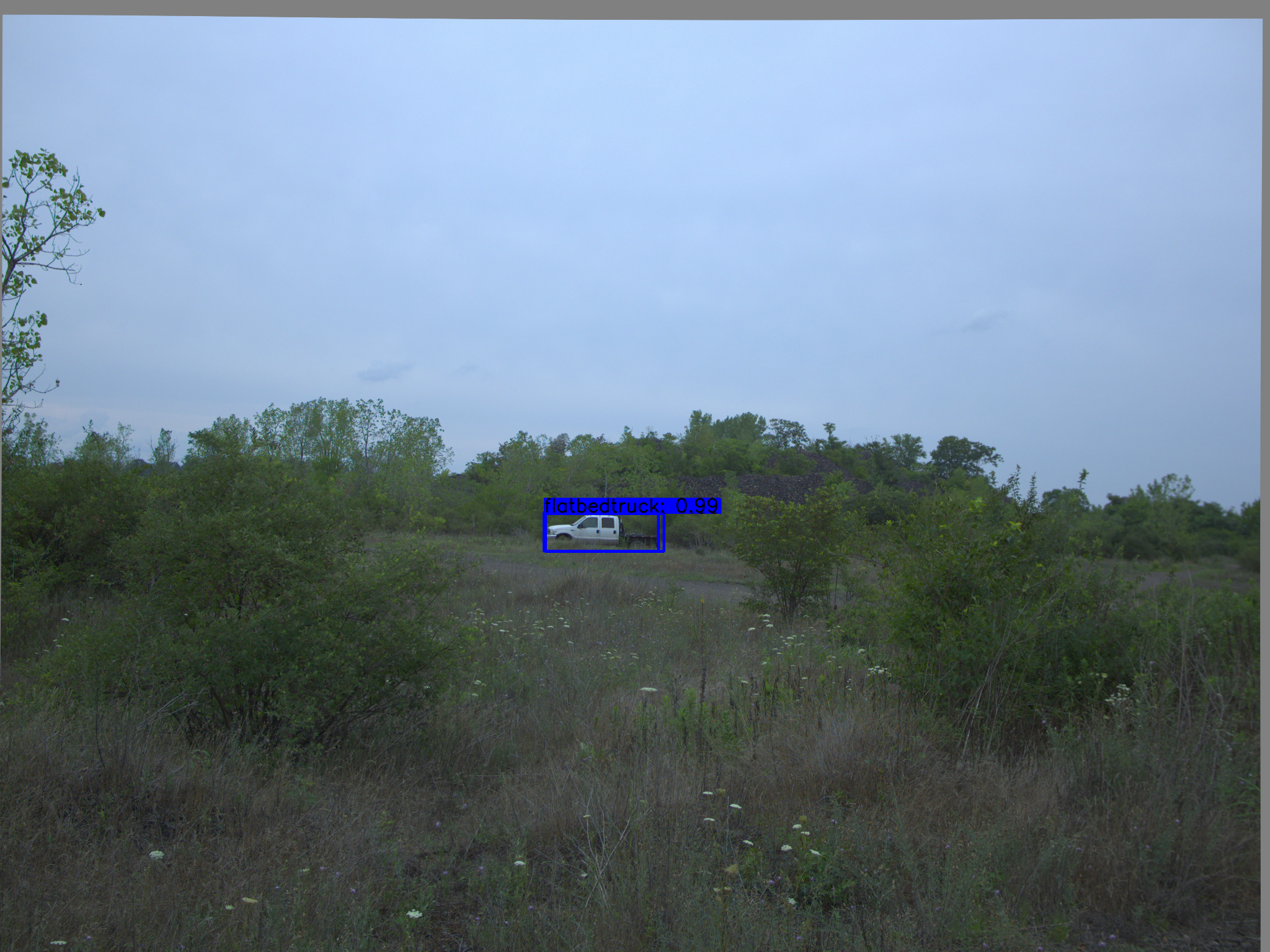}
     \end{subfigure}
     \begin{subfigure}{0.15\textwidth}
         \centering
         \includegraphics[width=\textwidth]{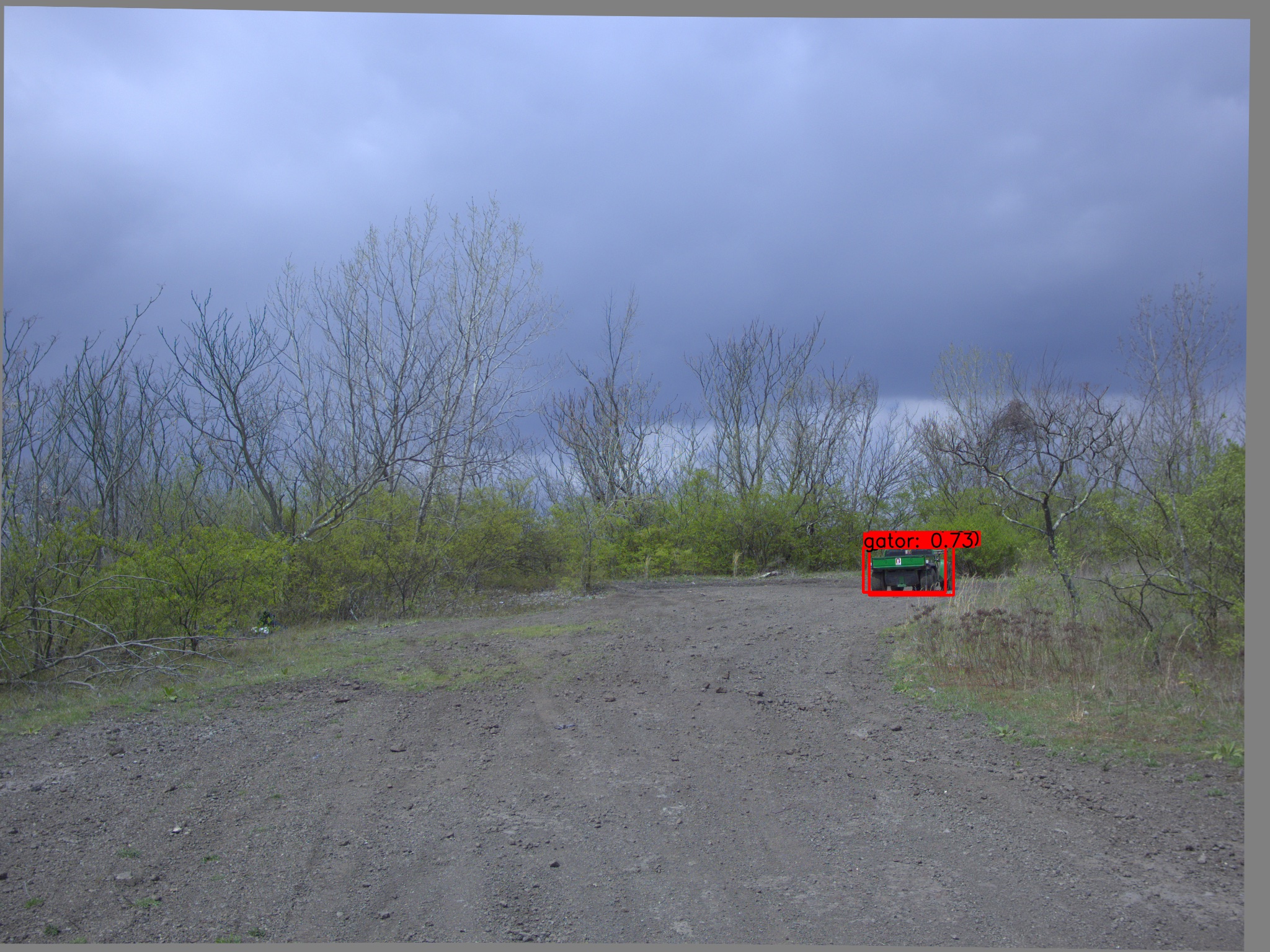}
     \end{subfigure}
     \begin{subfigure}{0.15\textwidth}
         \centering
         \includegraphics[width=\textwidth]{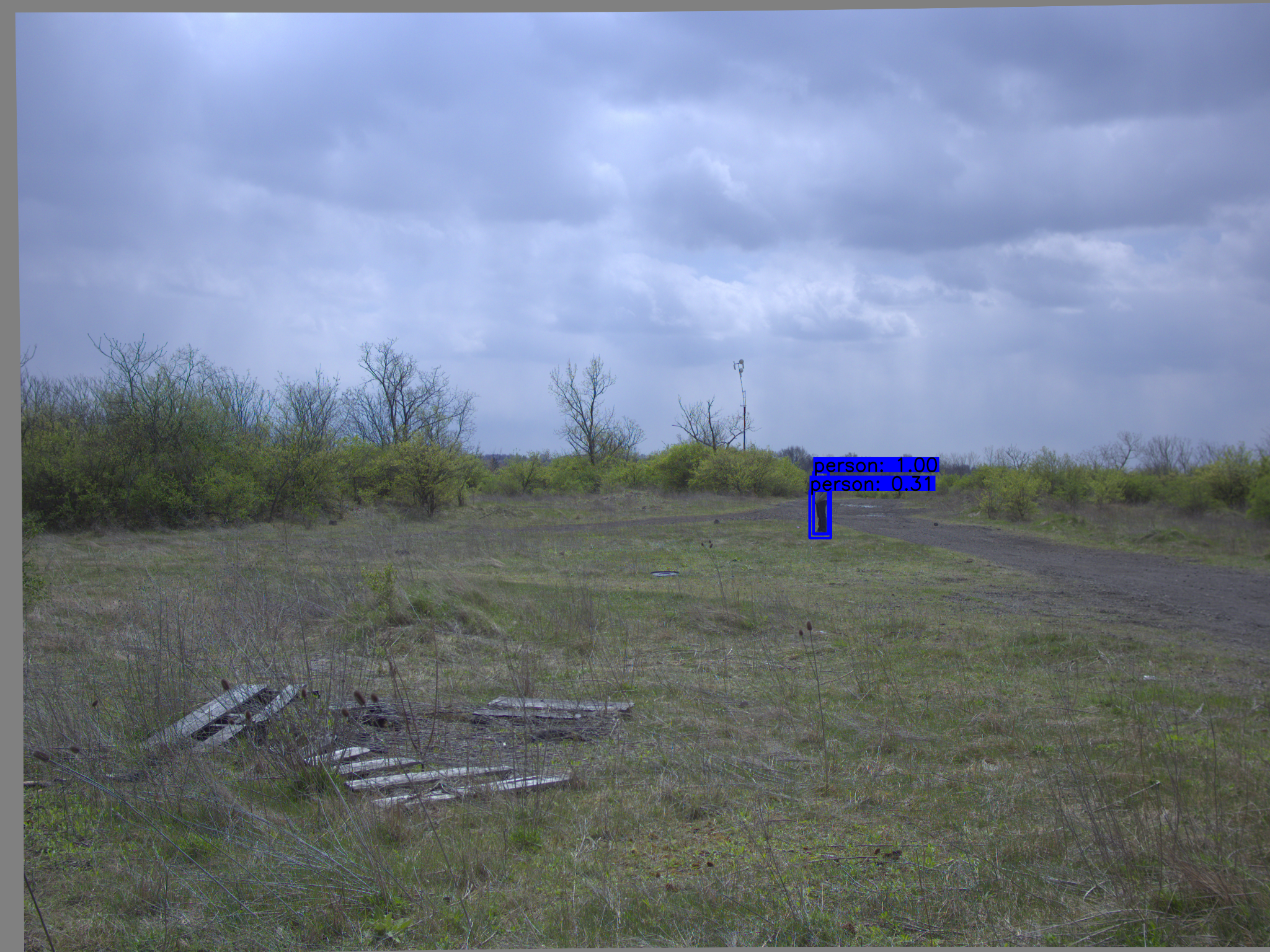}
     \end{subfigure}
     \centering
     \begin{subfigure}{0.15\textwidth}
         \centering
         \includegraphics[width=\textwidth]{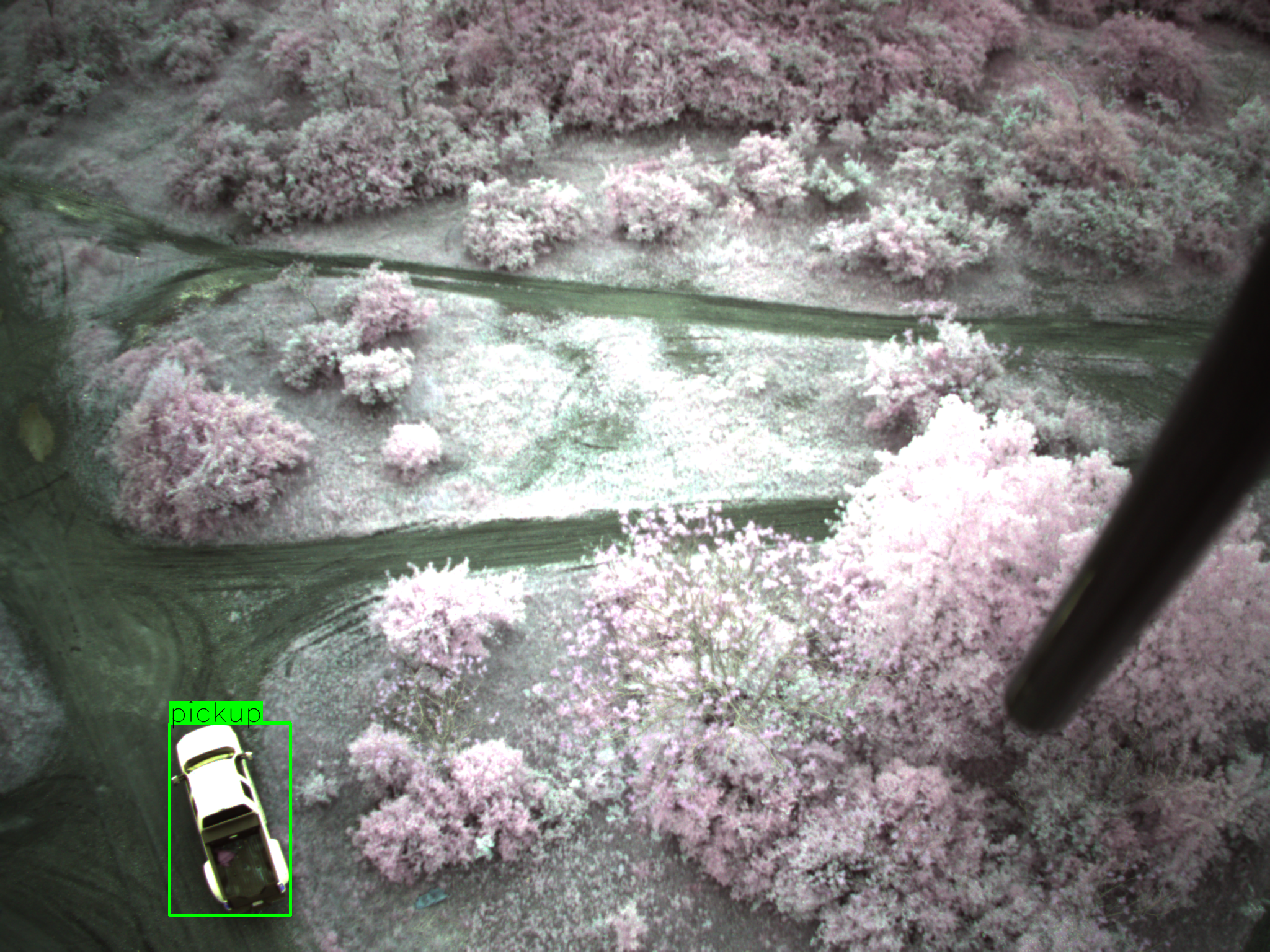}
         \caption{Pickup Truck}
     \end{subfigure}
     \begin{subfigure}{0.15\textwidth}
         \centering
         \includegraphics[width=\textwidth]{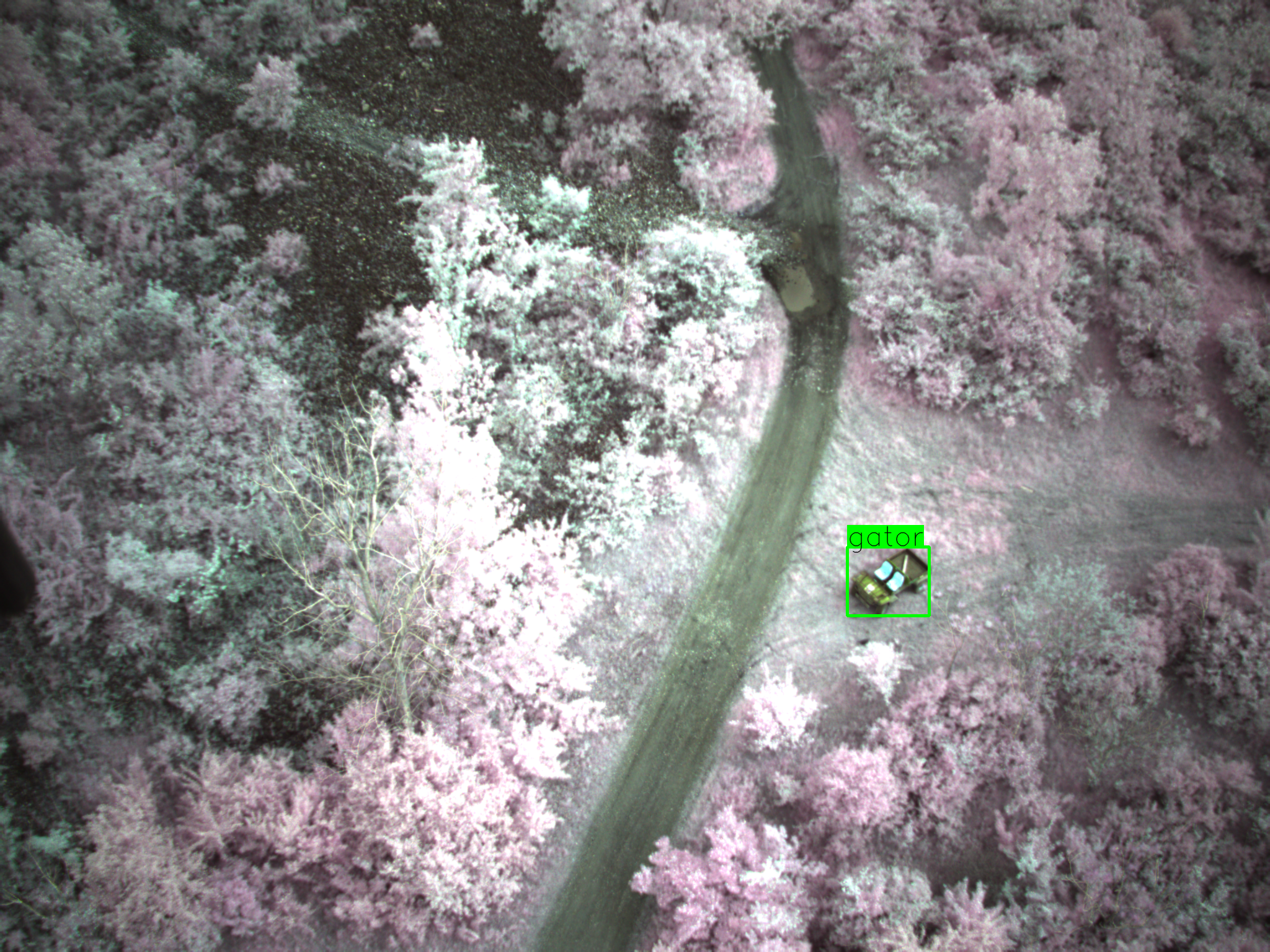}
         \caption{E-Gator}
     \end{subfigure}
     \begin{subfigure}{0.15\textwidth}
         \centering
         \includegraphics[width=\textwidth]{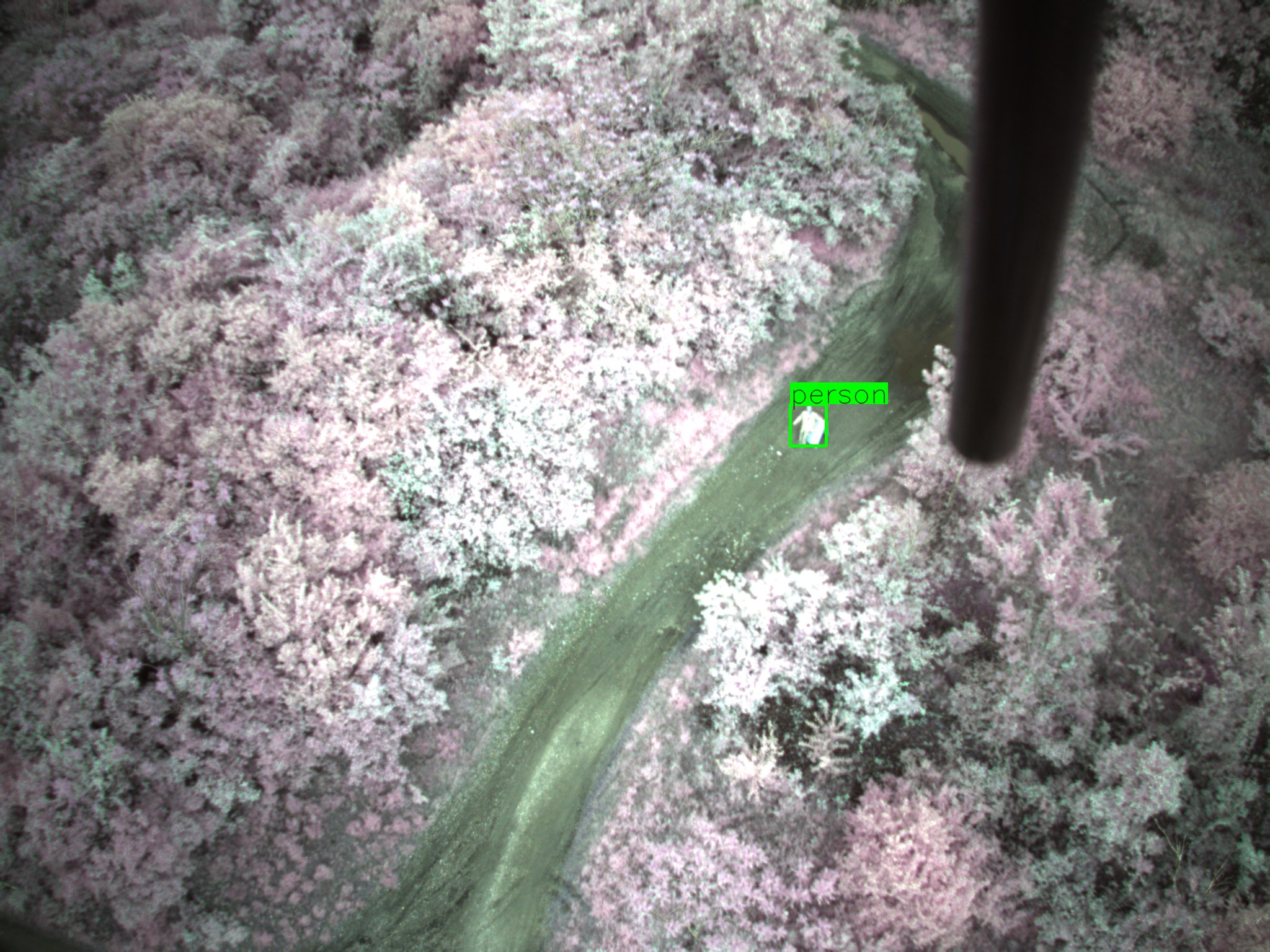}
         \caption{Person}
     \end{subfigure}
    \caption{Object detector: qualitative results. The first row comes from UGV detector and the second row comes from UAV detector.}
    \label{fig:det_quali}
\end{figure}

\subsection{Perception}
Fig. \ref{fig:det_quali} presents qualitative results for the UGV detector (first row) and UAV detector (second row). In our experiment, we detect three classes of objects: person, E-Gator, and pickup truck. The vegetation in the UAV pictures shows pink color because the UAV lens does not include any filtering and the camera is capturing the IR spectrum in addition to the RGB spectrum. Fig. \ref{fig:detbenchmark} shows the UGV detector benchmark. A single Nvidia RTX 6000 GPU is used to run 5 instances of the detector for the 5 RGB cameras in the sensor pod. The inference speed in Fig. \ref{fig:detbenchmark} refers to one instance. Fig. \ref{fig:gatormap} shows the UGV PR curve corresponding to 2048$\times$1536 input image resolution. We use this resolution for our field test.



\begin{figure}[h]
\centering\includegraphics[width=0.30\textwidth]{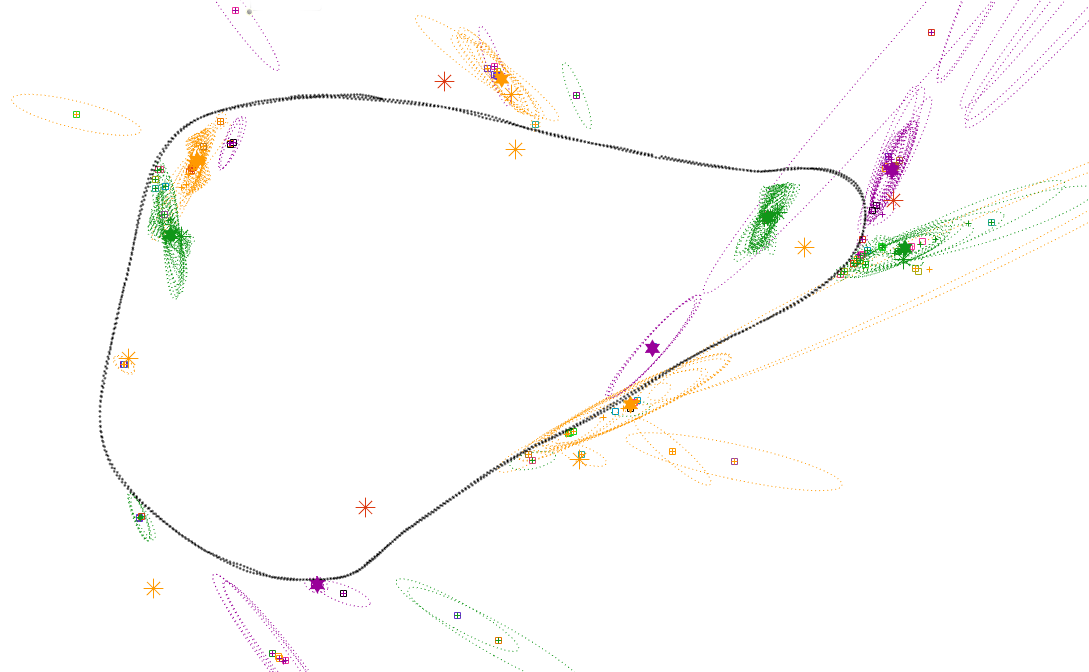}
\setlength{\belowcaptionskip}{-14pt}
\caption{Contact, Track, and vehicle positions. Dotted black points are UGV trajectory. Contacts are visualized as plus sign for their detected position and ellipses for their covariance. Track positions are shown as squares as they converge to their final position, represented by a six-sided star. Ground truth track position are indicated as asterisks. Colors of track squares represent their id, while colors of the contacts, final track positions, and ground truth positions represent their object classes.}
\label{fig:contact_covar}
\end{figure}

As explained in Section \ref{sec:ai}, we project 2D detections to 3D using pointcloud generated from stereo vision machine learning model. In order to achieve high precision, the stereo image pairs are well calibrated using a large 4ft by 8ft AprilTag board. Each UGV runs 5 stereo model instances on 2 Nvidia P6000 GPUs. The input images are rectified and undistorted. The UAV does not run stereo model because it detects objects in bird's eye view so it is straightforward to map the detection to global UTM coordinates.
Fig \ref{fig:contact_covar} visualizes the detected object covariance and final track position. 
We don't run tracker on UAV during our field tests.


\begin{figure}[tb]
\centering\includegraphics[width=0.48\textwidth]{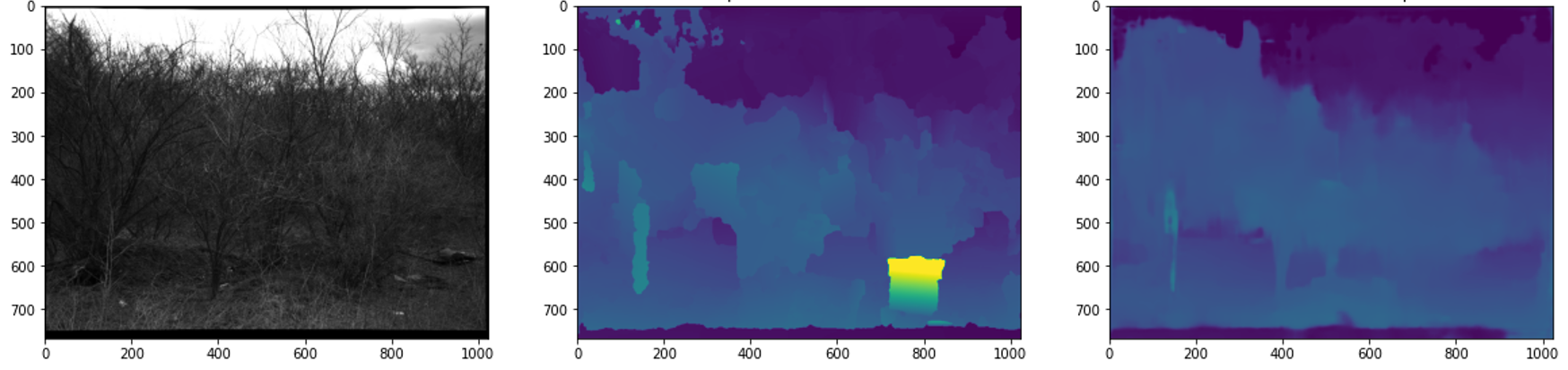}
\vspace{-12pt}
\setlength{\belowcaptionskip}{-14pt}
\caption{Stereo depth estimator: qualitative results. Left: NIR image from top camera. Middle: PRSM pseudo ground truth disparity map. Right: Predicted disparity map from the model.}
\label{fig:depth}
\end{figure}

\subsection{Navigation}
\begin{figure}[h]
\centering\includegraphics[width=0.3\textwidth]{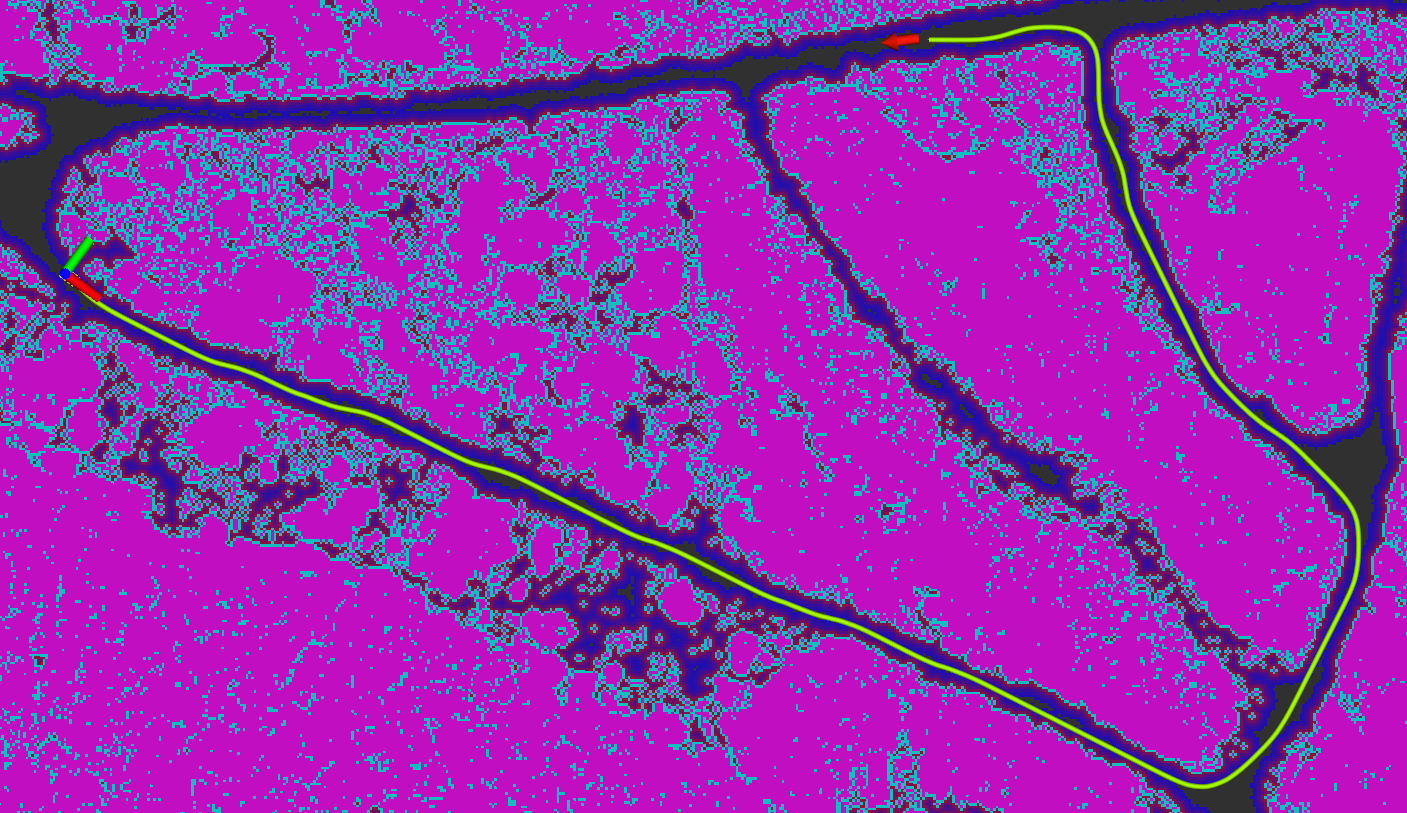}
\caption{An instance of the trajectory generated by the global planner between the current robot pose (perpendicular red-green lines) and the input goal point (red arrow) is shown as a green line.
}
\label{fig:global_plan}
\vspace{-12pt}
\end{figure}

\begin{figure}[ht]
\centering\includegraphics[width=0.3\textwidth]{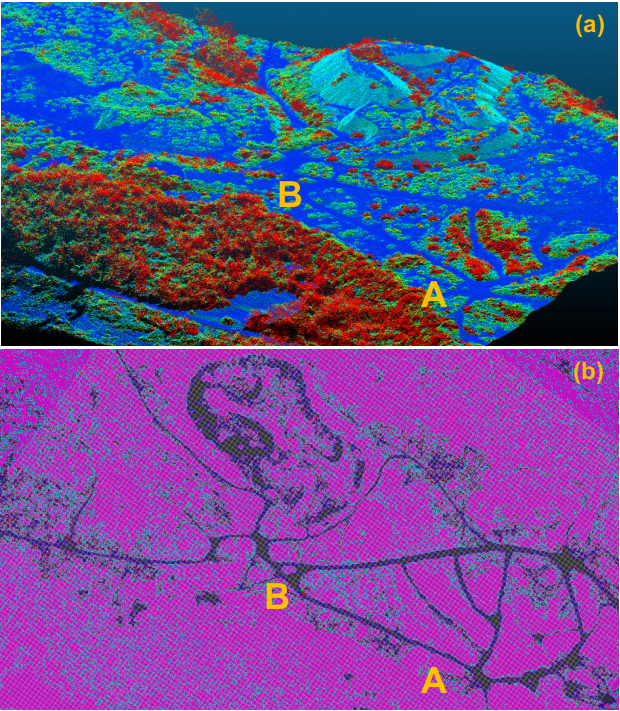}
\caption{Global map: The roughness of the terrain is calculated directly from a 3D point cloud (a). The color map indicates the terrain roughness: lowest (blue) to highest (red). The cells whose roughness is smaller than a certain threshold are projected onto a 2D cost-map (b). Markers A and B provide a reference for comparison.
}
\label{fig:gmap}
\end{figure}

As part of the field experiments, a sequence of GPS-referenced way-points were manually input to the navigation system. The ARA* global planner computed a global plan connecting the input way-points while optimizing for the cost of the trajectory as derived from the prior global cost-map (shown in Figure~\ref{fig:gmap}). An instance of the generated global plan (shown in green) between the current robot pose (seen as a perpendicular red-green line) and the goal way-point (represented as a red arrow) is illustrated in Figure~\ref{fig:global_plan}. A pure-pursuit local planner actuated the UGV to track the trajectory generated by the global planner. The pure-pursuit planner was tuned to track trajectories at a constant look-ahead distance of 8 meters and drove the UGV at a constant velocity of 3 m/s.

\begin{figure}[h]
\centering\includegraphics[width=0.3\textwidth]{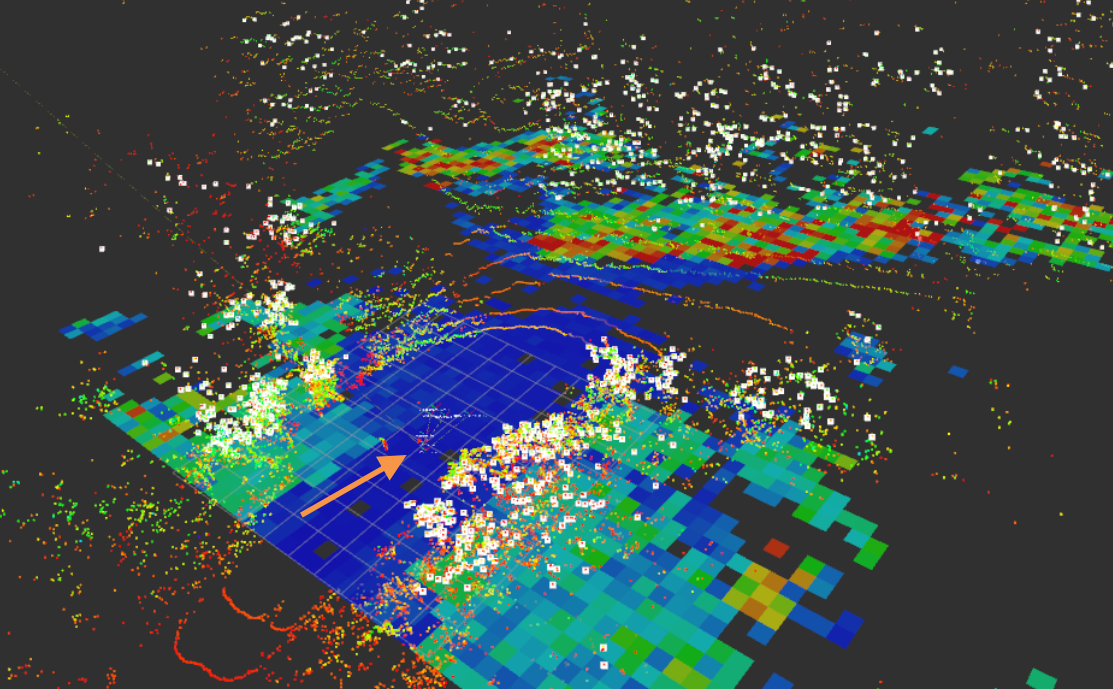}
\caption{Local cost-map: This example shows the raw LiDAR measurements (colored points). The ground elevation map is indicated as a colored cost-map (blue is low and red is high), and the white markers indicate the points above the ground plane, used to identify obstacles. The robot is located at the position indicated by the arrow.}
\label{fig:cost}
\vspace{-20pt}
\end{figure}

Fig. \ref{fig:cost} depicts an example of a local cost-map from our field test.
Fig. \ref{fig:gmap} shows the original point cloud model from where the terrain roughness is calculated, and the resulting map in 2D, which is used by the global planner.


\subsection{System Software}

\begin{figure}[!htb]
\centering\includegraphics[width=0.48\textwidth]{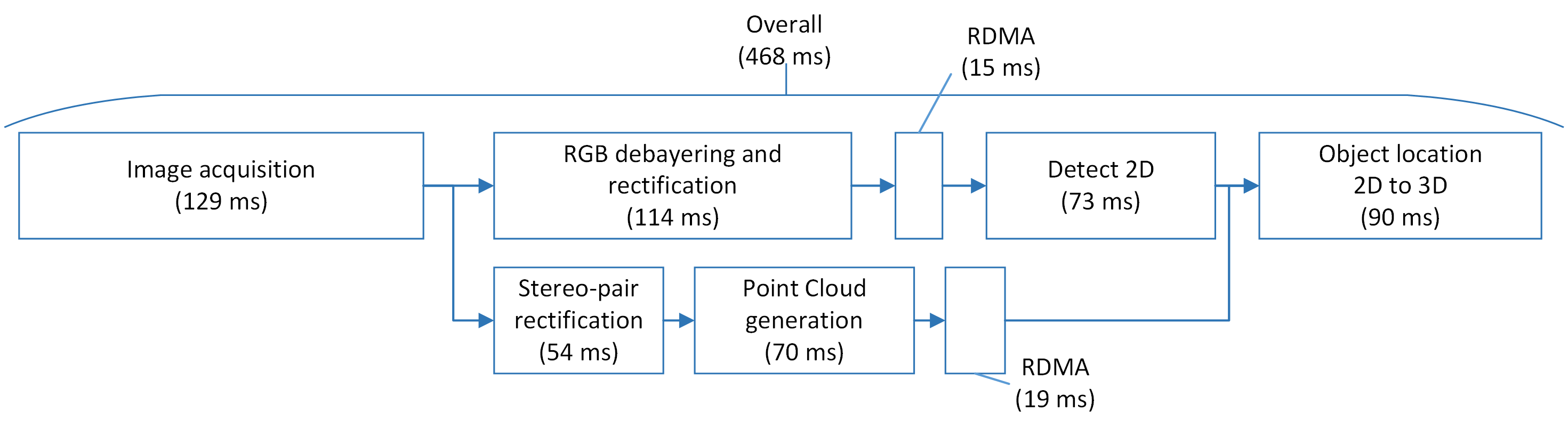}
\caption{Timeline of object geolocation. The width of the boxes are to scale.}
\label{fig:timeline}
\end{figure}

We measured different aspects of our on-board software. First, the end-to-end delay. This included calculating the difference between timestamps generated for each step where the images (and then point clouds) are passed in the pipeline. This starts when the image is triggered compared to when it is actually captured. Once it is captured, we look at when it reaches CPU memory, then when it reaches the pre-processing of the RGB/Thermal/NIR images on the I/O computers. For the NIR images, we look at the time it takes from the pre-processing the images to generating the point clouds on the GPU. Next, we calculate how long the RDMA transfer of the images and point clouds take. We also determine the duration of detecting objects. Finally, we calculate how long it takes to determine a 3D location of an object in the global frame from a 2D detection. The results of the tests are shown in Fig. \ref{fig:timeline}. As can be seen in the figure, processing of the RGB cameras and the stereo-pair are carried out in parallel, also image acquisition for the next image occurs while the current image is still be processed. This is our first implementation for the data transfer pipeline, and not all possible optimizations have been used.

\subsection{Field Testing}
\begin{figure}[ht!]
\centering\includegraphics[width=0.35\textwidth]{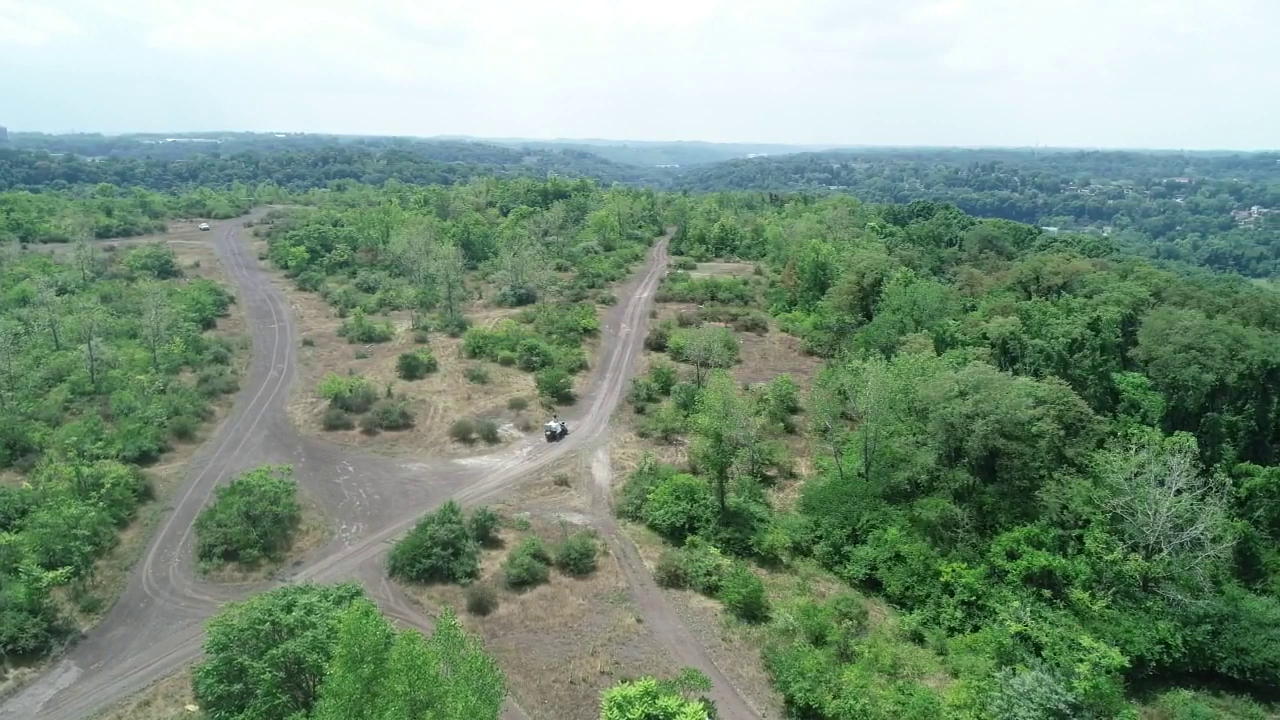}
\setlength{\belowcaptionskip}{-12pt}
\caption{Off campus test field in Pittsburgh, PA.}
\label{fig:gascola}
\end{figure}

Extensive field testing was performed to verify our system. The testing site used is approximately 100 acres and is located near CMU's main campus in Pittsburgh, Pennsylvania. An overhead view of the UGV performing a run at the test site can be see in Fig. \ref{fig:gascola}.
\begin{figure}[ht!]
\centering\includegraphics[width=0.35\textwidth]{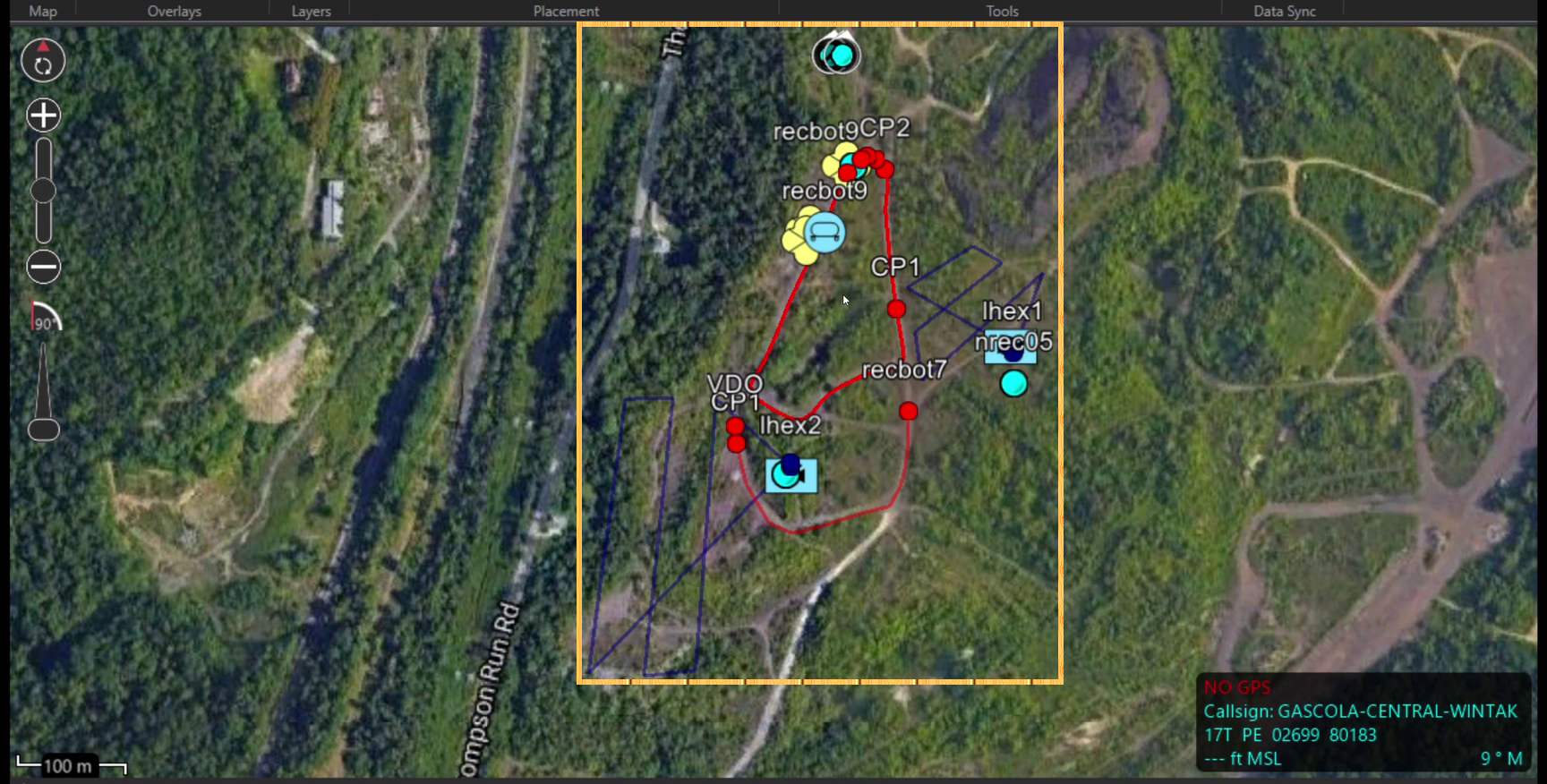}
\caption{COP interface with multiple robots and multiple detections. Yellow box showes the extent of the test site.}
\label{fig:COP}
\end{figure}

\section{Conclusion} \label{sec:conclusion}
Our UGV-UAV system is the foundation for multiple future research directions. First, 
the proposed system is a new platform to collect off-road data for related algorithmic research on perception and autonomy. 
Second, this system is a  foundation for future cooperative robotic system research, including: how to jointly optimize the hardware, on-board system software, and AI algorithms to make it be more cost effective and more robust to failures; how to coordinate the UAVs and UGVs in the system autonomously without centralized control. 
Third, the platform will be used to develop and test new functionalities such as visual odometry in GPS denied environments \cite{gpsdeny}, automatic way-point selection algorithms \cite{multiagent}, and RL-based development of system-level strategies.


%

\section*{Acknowledgment}
The authors would like to thank Bob Bittner, Anton Dugas, Edward Finely, and Jake Lammott from the field test team, Justin Decheck and Jim Ketterer our technicians, Ben Carter, Greg Paff, Graham Papciak, and Gary Russell from the machine shop, Jesse Holdaway and Jessica Kasemer from the project management team, Don Wilkinson from IT, Daniel Bittner, Zach DeCarolis, Krissy Guttendorf, Samuel Lisa, Patrick Pitman, and Josh Siniscal from the labeling team, Sean Bryan, Bob Ferguson, Jon Hamilton, David Hedger, Erik Kahn, Meghan Kokoski, Jason Kulk, Karl Muecke, Rich Pantaleo, Nishant Pol, Eric Sample, Eric Schmidt, Doug Soxman, Andre Sutanto, JP Tardif, and Brian Wallace from the engineering team, and Divam Gupta from CMU Auton Lab. We would also like to thank our academic partners including CMU DeLight Lab, TAMU Vehicle Systems \& Control Lab, Penn State University Applied Research Lab, and CMU Software Engineering Institute.


This project was funded by the US ARMY ACC-APG-RTP under  Contract  No. W911NF1820218.



\bibliographystyle{IEEEtran}
\bibliography{aidtr_year1_iros}

 \newcommand{\noop}[1]{}
\begin{thebibliography}{10}
\providecommand{\url}[1]{#1}
\csname url@samestyle\endcsname
\providecommand{\newblock}{\relax}
\providecommand{\bibinfo}[2]{#2}
\providecommand{\BIBentrySTDinterwordspacing}{\spaceskip=0pt\relax}
\providecommand{\BIBentryALTinterwordstretchfactor}{4}
\providecommand{\BIBentryALTinterwordspacing}{\spaceskip=\fontdimen2\font plus
\BIBentryALTinterwordstretchfactor\fontdimen3\font minus
  \fontdimen4\font\relax}
\providecommand{\BIBforeignlanguage}[2]{{%
\expandafter\ifx\csname l@#1\endcsname\relax
\typeout{** WARNING: IEEEtran.bst: No hyphenation pattern has been}%
\typeout{** loaded for the language `#1'. Using the pattern for}%
\typeout{** the default language instead.}%
\else
\language=\csname l@#1\endcsname
\fi
#2}}
\providecommand{\BIBdecl}{\relax}
\BIBdecl

\bibitem{thermal_sensor}
A.~{Ordóñez Müller} and A.~{Kroll}, ``Generating high fidelity 3-{D}
  thermograms with a handheld real-time thermal imaging system,'' \emph{IEEE
  Sensors Journal}, vol.~17, no.~3, pp. 774--783, 2017.

\bibitem{tokyo_multispectral}
K.~Takumi, K.~Watanabe, Q.~Ha, A.~Tejero-De-Pablos, Y.~Ushiku, and T.~Harada,
  ``Multispectral object detection for autonomous vehicles,'' in
  \emph{Proceedings of the on Thematic Workshops of ACM Multimedia 2017}, 2017,
  pp. 35--43.

\bibitem{Uav19}
T.~{Miki}, P.~{Khrapchenkov}, and K.~{Hori}, ``{UAV/UGV} autonomous
  cooperation: {UAV} assists {UGV} to climb a cliff by attaching a tether,'' in
  \emph{2019 International Conference on Robotics and Automation (ICRA)}, 2019,
  pp. 8041--8047.

\bibitem{petrovic2015can}
T.~Petrovic, T.~Haus, B.~Arbanas, M.~Orsag, and S.~Bogdan, ``Can {UAV} and
  {UGV} be best buddies? towards heterogeneous aerial-ground cooperative robot
  system for complex aerial manipulation tasks,'' in \emph{12th international
  conference on informatics in control, automation and robotics (ICINCO)},
  vol.~1.\hskip 1em plus 0.5em minus 0.4em\relax IEEE, 2015, pp. 238--245.

\bibitem{tanner2007switched}
H.~G. Tanner, ``Switched {UAV}-{UGV} cooperation scheme for target detection,''
  in \emph{Proceedings 2007 IEEE International Conference on Robotics and
  Automation}.\hskip 1em plus 0.5em minus 0.4em\relax IEEE, 2007, pp.
  3457--3462.

\bibitem{Balta2020}
\BIBentryALTinterwordspacing
H.~Balta, J.~Velagic, H.~Beglerovic, G.~De~Cubber, and B.~Siciliano, ``{3D}
  registration and integrated segmentation framework for heterogeneous unmanned
  robotic systems,'' \emph{Remote Sensing}, vol.~12, no.~10, p. 1608, May 2020.
  [Online]. Available: \url{http://dx.doi.org/10.3390/rs12101608}
\BIBentrySTDinterwordspacing

\bibitem{surmann20173d}
H.~Surmann, N.~Berninger, and R.~Worst, ``{3D} mapping for multi hybrid robot
  cooperation,'' in \emph{2017 IEEE/RSJ International Conference on Intelligent
  Robots and Systems (IROS)}.\hskip 1em plus 0.5em minus 0.4em\relax IEEE,
  2017, pp. 626--633.

\bibitem{kim2019uav}
P.~Kim, L.~C. Price, J.~Park, and Y.~K. Cho, ``{UAV-UGV} cooperative {3D}
  environmental mapping,'' in \emph{Computing in Civil Engineering 2019: Data,
  Sensing, and Analytics}.\hskip 1em plus 0.5em minus 0.4em\relax American
  Society of Civil Engineers Reston, VA, 2019, pp. 384--392.

\bibitem{hexy}
\url{https://vscl.tamu.edu/vehicles/little-hexy/}.

\bibitem{pixhawk}
\url{https://pixhawk.org/}.

\bibitem{yolov3}
J.~Redmon and A.~Farhadi, ``{YOLO}v3: An incremental improvement,''
  \emph{CoRR}, vol. abs/1804.02767, 2018.

\bibitem{multianchor}
W.~{Ke}, T.~{Zhang}, Z.~{Huang}, Q.~{Ye}, J.~{Liu}, and D.~{Huang}, ``Multiple
  anchor learning for visual object detection,'' in \emph{2020 IEEE/CVF
  Conference on Computer Vision and Pattern Recognition (CVPR)}, 2020, pp.
  10\,203--10\,212.

\bibitem{yang2019hsm}
G.~Yang, J.~Manela, M.~Happold, and D.~Ramanan, ``Hierarchical deep stereo
  matching on high-resolution images,'' in \emph{The IEEE Conference on
  Computer Vision and Pattern Recognition (CVPR)}, June 2019.

\bibitem{Dosovitskiy17}
A.~Dosovitskiy, G.~Ros, F.~Codevilla, A.~Lopez, and V.~Koltun, ``{CARLA}: {An}
  open urban driving simulator,'' in \emph{Proceedings of the 1st Annual
  Conference on Robot Learning}, 2017, pp. 1--16.

\bibitem{prsm}
C.~{Vogel}, K.~{Schindler}, and S.~{Roth}, ``Piecewise rigid scene flow,'' in
  \emph{IEEE International Conference on Computer Vision}, 2013.

\bibitem{unsupervise}
C.~Godard, O.~M. Aodha, and G.~J. Brostow, ``Unsupervised monocular depth
  estimation with left-right consistency,'' \emph{CoRR}, vol. abs/1609.03677,
  2016.

\bibitem{chiu2020probabilistic}
H.-k. Chiu, A.~Prioletti, J.~Li, and J.~Bohg, ``Probabilistic {3D} multi-object
  tracking for autonomous driving,'' \emph{arXiv preprint arXiv:2001.05673},
  2020.

\bibitem{arducopter}
\url{https://ardupilot.org/copter/}.

\bibitem{likhachev2004ara}
M.~Likhachev, G.~J. Gordon, and S.~Thrun, ``{ARA}*: Anytime {A}* with provable
  bounds on sub-optimality,'' in \emph{Advances in neural information
  processing systems}, 2004, pp. 767--774.

\bibitem{coulter1992implementation}
R.~C. Coulter, ``Implementation of the pure pursuit path tracking algorithm,''
  Carnegie-Mellon UNIV Pittsburgh PA Robotics INST, Tech. Rep., 1992.

\bibitem{snider2009automatic}
J.~M. Snider \emph{et~al.}, ``Automatic steering methods for autonomous
  automobile path tracking,'' \emph{Robotics Institute, Pittsburgh, PA, Tech.
  Rep. CMU-RITR-09-08}, 2009.

\bibitem{atak}
\url{https://en.wikipedia.org/wiki/Android\_Team\_Awareness\_Kit}.

\bibitem{gpsdeny}
J.~{Hsiung}, M.~{Hsiao}, E.~{Westman}, R.~{Valencia}, and M.~{Kaess},
  ``Information sparsification in visual-inertial odometry,'' in \emph{2018
  IEEE/RSJ International Conference on Intelligent Robots and Systems (IROS)},
  2018, pp. 1146--1153.

\bibitem{multiagent}
R.~Ghods, W.~J. Durkin, and J.~Schneider, ``Multi agent active search using
  realistic depth-aware noise model,'' \emph{IEEE Robotics and Automation
  Letters (RA-L)}, 2021.

\end{thebibliography}



%






\end{document}